\documentclass[11pt]{article}

\usepackage[letterpaper, margin=1in]{geometry}
\usepackage{graphicx}
\usepackage{amsmath,amssymb}
\usepackage[hidelinks]{hyperref}
\usepackage{xcolor}
\usepackage{booktabs}
\usepackage[super,sort&compress]{natbib}
\usepackage[utf8]{inputenc}
\usepackage[T1]{fontenc}
\usepackage{microtype}

\newcommand{\AUGpos}{\text{AUG}_{\text{pos}}}
\newcommand{\AUGnorm}{\text{AUG}_{\text{norm}}}
\newcommand{\Tstar}{T^{*}}

\title{Learning under noisy supervision is governed\\ by a feedback--truth gap}

\author{%
Elan Schonfeld\textsuperscript{1,*} \quad Elias Wisnia\textsuperscript{2} \\[0.8em]
{\small \textsuperscript{1}Department of Biology, Columbia University, New York, NY, USA} \\
{\small \textsuperscript{2}Department of Chemistry, Columbia University, New York, NY, USA} \\[0.3em]
{\small \textsuperscript{*}Corresponding author: elan.schonfeld@columbia.edu}
}

\date{}

\begin{document}

\maketitle

\begin{abstract}
When feedback is absorbed faster than task structure can be evaluated, the learner will favor feedback over truth. A two-timescale model shows this feedback--truth gap is inevitable whenever the two rates differ and vanishes only when they match. We test this prediction across neural networks trained with noisy labels (30 datasets, 2,700 runs), human probabilistic reversal learning ($N = 292$), and human reward/punishment learning with concurrent EEG ($N = 25$). In each system, truth is defined operationally: held-out labels, the objectively correct option, or the participant's pre-feedback expectation -- the only non-circular reference decodable from post-feedback EEG. The gap appeared universally but was regulated differently: dense networks accumulated it as memorization; sparse-residual scaffolding suppressed it; humans generated transient over-commitment that was actively recovered. Neural over-commitment (${\sim}0.04$--$0.10$) was amplified tenfold into behavioral commitment ($d = 3.3$--$3.9$). The gap is a fundamental constraint on learning under noisy supervision; its consequences depend on the regulation each system employs.
\end{abstract}

\section*{Introduction}

In many domains, learners do not receive truth as a direct instructional cue. Rather, learners update their internal representations of the world based upon feedback that only imperfectly reflects the truth. In machine learning, noisy labels can lead to memorization and poor generalizability\cite{arpit_closer_2017,zhang_understanding_2017,frenay_classification_2014,song_learning_2023}. In human decision-making, probabilistic outcomes can blur the distinction between contingency and chance, requiring the learner to act on noisy feedback\cite{daw_cortical_2006,sutton_reinforcement_2020}. While these settings seem quite different, both represent examples of noisy supervision. Several groups have argued for deeper integration between AI and neuroscience\cite{hassabis_neuroscience-inspired_2017,lake_building_2017,richards_deep_2019,marblestone_toward_2016}; however, we have lacked a common quantitative framework for understanding how noisy supervision affects learning processes in artificial and biological systems.

Here, we describe such a framework -- the feedback--truth gap -- the extent to which a learner's internal representation of the environment tracks feedback more than it tracks truth. We operationally define this term in three systems: for a neural network, the training-minus-validation accuracy difference as a function of epoch; for a human reversal task, the distance between feedback-aligned and truth-aligned responses to contingency reversals; and for an EEG experiment, the difference between post-feedback neural decoding of the feedback value and the participant's pre-feedback expectation (see Methods). Critically, in the EEG system the truth reference is the participant's prior expectation of correctness -- an operational proxy, not the latent task state -- because no non-circular reconstruction of objective environmental truth was decodable from the post-feedback signal (Extended Data Fig.~10). While the definitions vary across systems, they share the same formal properties: the learner revises its internal state based upon feedback before it becomes possible to resolve the discrepancy between the learner's representation and the true state of the environment. Memorization dynamics in neural networks have been studied extensively\cite{feldman_what_2020,toneva_empirical_2019}, and so have perseverative errors after contingency reversals in animals and humans\cite{cools_defining_2002,izquierdo_neural_2017,schoenbaum_new_2009,clark_neuropsychology_2004}. Computational models of human learning have focused on asymmetric learning rates for reward and punishment\cite{collins_how_2012,niv_neural_2012,dayan_reward_2002}. However, none of these previous studies have provided a necessary condition for when a learner will inevitably become over-committed to feedback, nor have they linked post-reversal over-commitment in humans to the generalization gap in machine learning.

Using a minimal two-timescale learner, we establish that a feedback--truth gap is inevitable when feedback is integrated on a faster timescale than the learner's evaluation of the task structure, and that the gap disappears only when the two timescales are equal (Methods; Fig.~2). We then empirically test this prediction across three independent systems -- neural networks trained with label noise (30 datasets, 2,700 runs), human probabilistic reversal learning ($N = 292$), and human reward/punishment learning with concurrent EEG ($N = 25$) -- using the same gap metrics ($\AUGpos$, $\Tstar$) across all three systems.

We found that the gap appeared in all three systems but was regulated in very different ways (Fig.~1). In dense neural networks, the gap grew progressively and became a permanent feature of the network's behavior -- memorization. In sparse-residual architectures and noise-robust training, the gap was reduced or suppressed (Fig.~4). In human reversal learning, the gap increased immediately following contingency reversals and was subsequently decreased through active behavioral regulation; the relationship between the size of the gap and the recovery time varied among individuals (Fig.~3). In EEG, post-feedback activity represented feedback valence and prior expectations, and individual differences in the neural gap predicted the amount of behavioral commitment each individual exhibited (Fig.~5). The same measurable quantity that indexed damage to learning in unregulated neural networks had the opposite implications for performance in regulated human learning.

\section*{Results}

\subsection*{The feedback--truth gap appears across learning systems}

We quantified the feedback--truth gap in three independent systems: artificial neural networks, human probabilistic reversal learners, and human reward/punishment learners with concurrent EEG (Fig.~1). The gap represents the same formal property of the learner's behavior -- feedback-aligned minus truth-aligned performance -- but was measured using each system's own notions of feedback and truth. The gap is therefore a common measurement construct, not a claim of common mechanism.

In neural networks trained on tabular data with 40\% symmetric label noise, training accuracy rose faster than validation accuracy, resulting in a persistent positive gap that grew throughout training (Fig.~1a). This gap resulted from progressive memorization of the noisy labels; the network tracked feedback -- including noise -- at the cost of generalizing to new, unseen data. No mechanism in either the training process or the network architecture prevented the growth of the gap. Across five benchmark datasets (Ionosphere, Glass, Sonar, WDBC, Vehicle) and six architectures, including label smoothing, $L_{2}$ regularization, and residual variants as controls, the gap was observed in greater than 90\% of model-noise combinations at 40\% label noise. The gap increased with the label noise rate and disappeared completely when the label noise rate was 0\% (Extended Data Figs.~1 and 2). The unregulated dense neural network thus provides the reference prediction of the two-timescale model, while the control conditions verify that the gap persists under conventional remediation and is not specific to any particular architecture.

In human probabilistic reversal learning ($N = 292$, ages 8--30; ref.~11), participants chose between two options with 75\%/25\% reward probabilities that reversed periodically (empirical noise rate: 17.1\%). We calculated the same gap metric -- feedback-aligned minus truth-aligned choice accuracy -- on a rolling trial-by-trial basis, synchronized with the contingency reversals. Following each reversal, feedback-aligned accuracy recovered more quickly than truth-aligned accuracy, resulting in a transient positive gap that peaked within approximately five trials and resolved over 15--20 trials (Fig.~1b). This signal was extremely reliable ($\AUGpos = 0.049 \pm 0.032$, $t(291) = 26.4$, $d = 1.55$, $p < 10^{-75}$) and nearly universal: all 292 subjects demonstrated positive over-commitment following a reversal, and 99.3\% eventually regained stable truth-aligned responding (mean $\Tstar = 9.9$ trials). Unlike in networks, the human gap resolves: active behavioral regulation closes it.

In concurrent EEG recordings during reward and punishment learning ($N = 25$; ref.~12), we decoded post-feedback feedback valence and each subject's pre-feedback expectation from the post-feedback EEG (Methods). Only one of four candidate truth references -- the participant's prior expectation of correctness -- was reliably decodable in this window; objective reconstructions of the task state based on Bayesian and reinforcement-learning models were not decodable (Extended Data Fig.~10). The expectation-truth decoder performed above chance in both tasks (REW: AUROC $= 0.541$, $t(23) = 2.43$, $p = 0.012$; PUN: AUROC $= 0.545$, $t(24) = 3.84$, $p < 10^{-3}$). The primary endpoint is not individual decoder accuracy but the trial-by-trial gap between the two decoders -- how much more confidently the EEG signal represents feedback valence than expected accuracy. The neural gap was positive in all subjects included in the analysis (reward: $N = 24$; punishment: $N = 25$; Extended Data Fig.~3), and feedback-dominant signals in the EEG diminished over trials while expectation-related signals followed feedback (Fig.~1c), consistent with progressive adaptation of the neural response.

Fig.~1 thus illustrates a hierarchy of regulation. Over-commitment to feedback occurred in all three systems, but the progression of the gap through time differed. Dense networks drifted further from truth; human choices returned to truth; and EEG signals adapted at an intermediate timescale. Two questions follow: why does the gap occur at all, and what determines whether the system regulates it?

\subsection*{The gap is inevitable once timescales differ}

We analyzed the minimal two-timescale learner to determine why the gap occurs so uniformly in our experiments. The model uses a fast feedback-integrating channel (rate $\alpha_{\text{fast}}$) and a slower truth-integrating channel (rate $\alpha_{\text{slow}}$) (Fig.~2; Methods). Noise in the feedback causes any change in the underlying state to expose the difference in timescales: the fast channel moves before the slow channel, producing a temporary separation of feedback tracking and truth tracking.

Using a closed-form solution (Supplementary Derivation), the gap after a state change of size $\Delta$ at $t = 0$ is:
\[
\text{gap}(t) = \Delta \cdot \bigl(e^{-\alpha_{\text{slow}} \cdot t} - e^{-\alpha_{\text{fast}} \cdot t}\bigr)
\]
which is positive for all $t > 0$ whenever the two rates differ (Fig.~2d). The peak gap magnitude depends on the timescale ratio $r = \alpha_{\text{fast}} / \alpha_{\text{slow}}$: it approaches $\Delta$ as $r$ grows large and shrinks to zero as $r$ approaches 1. If the two timescales match ($r = 1$), the gap is identically zero regardless of the noise level. In simulation, at $r = 1$ with 20\% noise, the mean peak gap stayed below $10^{-5}$ (Fig.~2d); across 40 timescale ratios the correspondence between theory and simulation gave $R^{2} > 0.99$.

Based on these findings, we created a phase diagram of the two-timescale learner with respect to timescale ratio and noise level (Extended Data Fig.~2c). The only no-gap boundary in timescale space is when the two channels operate at the same rate ($r = 1$); simulations confirm the predicted gap sizes throughout the entire parameter range. All three empirical systems can be located on this diagram: dense networks fall in the high-ratio, high-noise region ($r \approx 10$--20), explaining their large persistent gaps; humans occupy a moderate-ratio regime ($r \approx 3$--5) where regulation can close the gap; sparse-residual networks have reduced effective ratios and correspondingly smaller gaps. All systems examined empirically demonstrated that removing label noise collapsed the gap and eliminated the advantage of sparse bottlenecks (Extended Data Fig.~5).

This result describes what must occur based on the two-timescale structure; it does not describe how any individual system implements learning. Once the feedback is noisy and is integrated faster than truth, the only way to prevent transient over-commitment is to delay, suppress, or compensate the gap through downstream regulation. Because the gap is deterministic given the timescale ratio, it is predictable from early data: the first 15 epochs of network training suffice to classify the eventual memorization regime (epochs 35+) with AUC $= 0.96$, and early-to-late prediction also works in both human paradigms (Extended Data Fig.~4).

\subsection*{The significance of the gap depends upon regulation}

Although the same gap construct appeared in every system examined, the relationship of the gap to performance differed qualitatively among the systems (Fig.~3). The significance of the gap is determined entirely by whether and how the system regulates it.

In neural networks under synthetic label noise, the cumulative gap magnitude ($\AUGnorm$) was a strong negative predictor of validation accuracy across 150 dataset--model pairs ($\beta = -0.641$, $p < 10^{-6}$, fixed-effects regression; Fig.~3a). Higher memorization resulted in lower generalization; the gap was detrimental. Under natural noise (crowd-labelled CIFAR-10N), this relationship weakened as model rankings compressed, and onset timing $\Tstar$ became the informative metric ($r = 0.672$, $p = 0.048$; Extended Data Fig.~2), demonstrating regime dependence in which metrics carry diagnostic information.

The relationship was reversed in human reversal learning. Subjects who exhibited larger transient gaps after contingency reversals also exhibited faster subsequent recovery of truth-aligned responding (Fig.~3b). The gap did not indicate damage; instead, it indicated the extent to which the subject engaged with the changed contingency, and a larger initial discrepancy between feedback tracking and truth tracking was associated with more rapid realignment. This does not imply that the gap was beneficial; rather, it implies that the downstream effects of the gap depended on whether regulatory mechanisms existed to close it. This reversal (negative gap--performance relationship in networks, positive gap--recovery relationship in humans) is the central dissociation: the same measurable quantity had opposite implications for performance depending on regulatory capacity.

The EEG data identified a neural basis for this dissociation. The neural gap was spatially organized, with feedback-dominant components at frontal electrodes and expectation-dominant components at posterior sites (Extended Data Fig.~3). Instead of a uniform scalar measure, the post-feedback signal separated evaluative and expectation-related components (Fig.~3c), consistent with the established role of medial frontal cortex in feedback monitoring\cite{cavanagh_frontal_2014,ridderinkhof_role_2004} and with the theta, P300, and ERN literatures on outcome evaluation\cite{polich_updating_2007,holroyd_neural_2002,walsh_learning_2012}.

The results thus defined three regulatory regimes. Dense networks lacked regulation, and the gap accumulated without restriction, directly indexing damage. Additional noise-robust methods (co-teaching, forward loss correction) suppressed but did not abolish the gap (Extended Data Table~1). Architectural constraints limited the ability to overfit the feedback, thereby suppressing the gap. Humans exhibited transient gaps and then recovered through behavioral adjustment. Although the phenomenon occurred in every system, the nature of the regulatory response determined the outcome.

\subsection*{Regulation is tunable and dissociates pressure from recovery}

Two complementary analyses assessed whether the regulation of the gap could be systematically controlled: architectural intervention in neural networks and cross-level integration in human learners (Figs.~4--5).

The strength of a sparse-residual architectural bottleneck was varied ($\alpha \in \{0.1, 0.25, 0.5, 1.0\}$) across five datasets with 40\% label noise (10 seeds per configuration). Both $\AUGnorm$ and $\Tstar$ increased monotonically with $\alpha$ (Spearman $\rho = 1.0$ for both metrics; Fig.~4a--b), demonstrating graded causal control over the memorization dynamics. At the dataset level, this monotonic suppression held in all datasets with measurable gaps ($\AUGnorm > 0.005$; $\rho = 1.0$ in 3/3), whereas datasets at the noise floor demonstrated no systematic trends because there was no gap to suppress (Extended Data Fig.~5). However, test accuracy did not follow this monotonic pattern: it peaked at $\alpha = 0.1$--$0.25$ and decreased at higher values (Fig.~4c). This non-monotonic dissociation demonstrated a capacity--memorization trade-off in which configurations that maximized suppression of the gap did not maximize generalization. (For comparisons with label smoothing, weight decay, and loss-correction methods\cite{patrini_making_2017,natarajan_learning_2013}, see Methods and Supplementary Information.) Additional validation across 30 OpenML datasets confirmed that sparse bottlenecks improved accuracy on 93\% of datasets at 40\% noise but provided no reliable advantage at 0\% noise (Extended Data Fig.~5; Extended Data Table~2). Suppression is noise-dependent. Sparse-residual architectures served as a causal probe for graded, monotonic control over the gap, not as a proposed method for practical noise robustness.

Behavioral over-commitment to feedback was large and consistent in the EEG dataset ($N = 25$; ref.~12). All 25 subjects in both reward and punishment conditions showed positive over-commitment: reward $t(24) = 16.6$, $d = 3.32$, $p < 10^{-14}$; punishment $t(24) = 19.6$, $d = 3.92$, $p < 10^{-15}$ (Fig.~5a). Decomposition into win-stay (${\sim}0.80$--$0.83$) and lose-shift (${\sim}0.20$--$0.23$) components demonstrated that over-commitment was largely due to high persistence in rewarding actions, not failure to switch after negative outcomes. The concurrent neural gap, as assessed by EEG decoders, was considerably smaller (neural $\AUGpos \approx 0.04$--$0.10$; Fig.~5b), demonstrating approximately tenfold amplification from neural representation to behavioral strategy.

The neural gap was highly reproducible (split-half correlation between even and odd trials: $r = 0.35$, $p < 10^{-5}$ for reward; $r = 0.45$, $p < 10^{-8}$ for punishment) and decreased from early to late trials in the punishment condition ($d = 0.82$, $p < 0.001$; Fig.~3d), suggesting that the neural gap tracks a learning-related process rather than noise or drift (Extended Data Fig.~3). At the individual level, subjects with larger neural gaps also exhibited stronger behavioral commitment (reward: $\rho = 0.43$, $p = 0.036$; punishment: $\rho = 0.42$, $p = 0.042$; Fig.~5c). Neither EEG channel significantly predicts single-trial behavioral choice; rather, both channels appear to provide noisy measures of the same slow-changing, task-dependent learning state. The amplification is concordant across levels and across individuals (see Extended Data Fig.~6 for further individual-differences analyses, Extended Data Fig.~9 for robustness to the class-balance exclusion threshold, and Extended Data Table~4 for reinforcement-learning parameter associations).

The gap is suppressed differently in each system. Architectural bottlenecks suppress the gap by reducing the capacity of the network. Human brains employ a dynamic process that amplifies modest neural over-representation into large behavioral commitments and then corrects them. While the gap exists in all three systems, the systems differ in how they regulate it.

\section*{Discussion}

The feedback--truth gap is inherently mathematical: when a learning system integrates feedback before it can verify the true state of affairs, over-commitment to the feedback is inevitable. This gap was consistently evident in all of the systems evaluated. More importantly, the critical finding is that systems differ categorically in how they respond. Networks without regulation generate the gap as damage; architectural constraints produce static suppression; and humans generate active recovery, with concordant amplification at the neural and behavioral levels. This taxonomy -- unregulated, suppressed, recovered -- provides a conceptual framework for evaluating how different systems manage a common computational constraint.

\textbf{Scope.} We do not contend that humans learn in the same way as artificial neural networks. The shared aspect of the research is a framework for measurement -- the feedback--truth gap -- applied to the same formal manipulation (perturbing the agreement between feedback and ground truth at each learning step) across substrates. The gap is an inherent consequence of timescale mismatch, not something we advocate as adaptive or optimal. The sparse-residual architectures serve as causal probes; the human neural evidence is correlational. In the machine-learning studies, the architectural manipulations serve as controlled probes rather than proposed general-purpose noise-resistant methods; topology ablations indicate that the sparse-residual scaffolding, not the graph structure itself, suppresses the gap (Extended Data Table~3).

\textbf{Shared framework, not shared mechanism.} Convergence across substrates represents shared measurement, not shared mechanism. The same quantitative metrics ($\AUGpos$, $\Tstar$, commitment indices) capture gap dynamics in network training curves, human reversal learning, and reward/punishment strategies. The formal structure is identical across domains: noisy supervision, time-varying commitment, and eventual alignment with truth. However, the substrate-specific process that generates and regulates the gap is different in each case, and convergent measurement does not imply convergent implementation.

\textbf{Regulation as the differentiator.} What converges is the principle that raw gap pressure is modulated before the gap is expressed behaviorally. In networks, architectural regularization delays and suppresses the gap. In humans, neural over-representation ($\rho = 0.43$ in reward, $0.42$ in punishment; both $p < 0.05$) is amplified roughly tenfold into behavioral over-commitment. This concordance holds at subject and block timescales; at the single-trial level, both neural and behavioral signals provide noisy estimates of a shared, slow-changing, task-dependent learning state. In networks the regulatory mechanisms are weight decay and capacity constraints. In humans, prefrontal control circuits are the natural candidate, and dopaminergic signaling -- which modulates learning from prediction errors\cite{schultz_neural_1997,schultz_dopamine_2016} and differentially shapes reward versus punishment learning\cite{frank_by_2004,frank_mechanistic_2006,cools_dopaminergic_2006} -- is a plausible substrate for the amplification we observe.

\textbf{EEG dissociation.} EEG provided a distinct dissociation: post-feedback features decode feedback valence and the participant's pre-feedback expectation, but do not reconstruct the latent task state (Extended Data Fig.~10). Post-feedback ERPs, especially the P300, are responsive to surprise (outcome minus expectation), resulting in a residue of prior belief rather than a direct representation of the environment. Thus, the neural feedback--truth gap represents the balance between outcome evaluation and expectation. By the time this balance is reached, the prior beliefs are already partially regulated; feedback and expectation representations are nearly equalized, so the much larger behavioral commitment is likely mediated by downstream control rather than simply scaled versions of the same neural signal.

\textbf{Regime dependence.} Neither the gap nor its diagnostic value is uniform across conditions. In networks, $\AUGnorm$ predicts accuracy under synthetic noise but loses discrimination under natural noise, where $\Tstar$ becomes informative instead. In humans, the neural gap is modest ($d \approx 1.0$) but is amplified manyfold at the behavioral level ($d = 3.3$--$3.9$), with significant cross-modal correlations in both tasks ($\rho = 0.43$, $0.42$; $p < 0.05$). The phase diagram (Extended Data Fig.~2c) captures this continuous variation: gap magnitude scales with noise rate and timescale ratio and vanishes at specific boundaries. No single metric is universally diagnostic; the informative metric depends on the noise regime. The capacity--memorization tradeoff connects to the double-descent literature\cite{belkin_reconciling_2019}, mathematical theories of learning dynamics\cite{saxe_mathematical_2019}, and work on architecture and generalization\cite{neyshabur_exploring_2017}. Episodic memory\cite{gershman_reinforcement_2017} may also play a role, particularly where long delays separate feedback from truth verification.

This study has several limitations. Primarily, the human neural evidence is correlational. An exploratory pharmacological reanalysis is consistent with dopaminergic modulation of gap regulation: in a double-blind crossover ($N = 27$; \cite{cavanagh_conflict_2014}), the D2/D3 agonist cabergoline flattened within-session gap regulation relative to placebo ($t(26) = 2.31$, $d = 0.44$, one-sided $p = 0.015$; Wilcoxon $p = 0.016$; Extended Data Fig.~7), and cumulative gap level showed a directionally consistent reduction ($d = -0.32$, one-sided $p = 0.053$). However, this result is preliminary; proper causal tests will require stimulation or lesion studies. A developmental analysis of the reversal-learning dataset ($N = 292$; Extended Data Fig.~8) indicates a shift from more flexibility in children to more persistent use of feedback in adults, but future studies should assess these relationships in more diverse populations. The human samples are Western adults performing laboratory tasks from public repositories; generalizability to naturalistic or developmental settings is untested. The two-timescale model determines when the gap should appear, not how specific systems create or manage it; attention, working memory, and normalization dynamics all modulate the gap in ways that timescale separation alone does not capture. Expanding the framework to reinforcement-learning agents, developmental populations where prefrontal regulation is still maturing\cite{casey_beyond_2015,somerville_developmental_2010,hartley_neuroscience_2015}, and clinical groups with abnormal feedback processing\cite{waltz_probabilistic_2007,peterson_probabilistic_2009,maia_reinforcement_2011,murray_substantia_2008,waltz_patients_2009} will provide evidence for the framework's broader utility.

The gap between feedback and truth should arise wherever there is noisy feedback and latent truth: scientific inference, social learning\cite{rendell_why_2010}, organizational adaptation, cultural transmission\cite{richerson_culture_1985}. The structure echoes foundational ideas in associative learning\cite{rescorla_theory_1972,siegel_widespread_1996}. Understanding what regulates the gap -- across architectures, development, and pathology -- may provide insight into developing both robust machine learning and human adaptive behaviors.

\section*{Methods}

\subsection*{Two-timescale learner model}

The simplest model exhibiting a feedback--truth gap tracks a binary environment via two exponential-moving-average channels with update rates $\alpha_{\text{fast}}$ and $\alpha_{\text{slow}}$ ($\alpha_{\text{fast}} > \alpha_{\text{slow}} > 0$). The observed feedback is noisy: on each trial, the observed signal matches the true state with probability $1 - \varepsilon$. After a state reversal of magnitude $\Delta$ at $t = 0$, the gap between the fast and slow estimates is $\text{gap}(t) = \Delta \cdot (e^{-\alpha_{\text{slow}} \cdot t} - e^{-\alpha_{\text{fast}} \cdot t})$, strictly positive for all $t > 0$ (Supplementary Derivation), with peak magnitude $\text{gap}_{\max} = \Delta \cdot r^{-1/(r-1)} \cdot (1 - r^{-1})$ where $r = \alpha_{\text{fast}} / \alpha_{\text{slow}}$. Simulations used 20 seeds per parameter setting with $\alpha_{\text{slow}} = 0.02$, noise $\in \{0.1, 0.2, 0.4\}$, and $r$ ranging from 1 to 63. Phase diagrams were constructed from $20 \times 20$ grids of noise rate $\times$ timescale ratio.

\subsection*{Machine learning experiments}

Five tabular datasets from UCI/OpenML (Ionosphere, Glass, Sonar, WDBC, Vehicle) were selected because memorization is clearly observable within 100 epochs at 40\% symmetric label noise. An extended benchmark of 30 OpenML datasets across three noise rates, three noise-type configurations, and ten seeds (2,700 runs total) confirmed generality. CIFAR-10 with synthetic noise and CIFAR-10N (human annotator noise) served as boundary cases.

Fully connected networks provided the dense baseline. The sparse-residual variant combines an identity shortcut with a fixed sparse bottleneck wired as an expander graph, reducing the dense layer's multiply-accumulate operations to roughly 22\%. Bottleneck strength $\alpha \in \{0.1, 0.25, 0.5, 1.0\}$ controls the capacity of the sparse branch. Three control architectures isolate specific ingredients: dense with label smoothing (Dense+LS), dense with a residual shortcut but no sparsity (Dense+Residual), and dense with strong $L_{2}$ (Dense+StrongReg).

Two metrics quantify the gap. $\AUGnorm$ is the normalized area under the positive portion of the training-minus-validation accuracy curve; $\Tstar$ is the first epoch at which the gap exceeds $\tau = 0.05$ for three consecutive epochs. Both were computed per random seed and averaged. Statistical inference relied on fixed-effects regression across 150 dataset--model pairs, Spearman correlations for monotonicity, and 95\% confidence intervals throughout. Each configuration had 10 seeds in the Full stage.

\subsection*{Human probabilistic reversal learning}

Trial-level behavioral data from 292 participants (ages 8--30) who performed a probabilistic reversal learning task\cite{eckstein_reinforcement_2022} were obtained from a publicly accessible repository (OSF; \url{https://osf.io/7wuh4/}). Participants played a two-armed bandit with 75\%/25\% reward probabilities that reversed periodically (4--9 reversals per session, 125--131 trials total). The empirical noise rate was 17.1\%.

Rolling accuracy was computed over an 8-trial window independently for truth-aligned and feedback-aligned responses; the gap is the difference (feedback minus truth), time-locked to each reversal. $\AUGpos$ is the baseline-corrected area under the positive portion of this curve from trials 0 to 25 post-reversal. $\Tstar$ is the first post-reversal trial at which truth accuracy exceeded 65\% for 15 consecutive trials. $\AUGpos$ was tested against zero with a one-sample $t$-test, reporting Cohen's $d$ and bootstrap 95\% CIs.

\subsection*{Human reward/punishment learning with EEG}

Behavioral and 32-channel EEG data from 26 healthy adults who performed separate reward and punishment probabilistic learning tasks\cite{stolz_reward_2022} were obtained from OpenNeuro (ds004295). One subject was dropped owing to an EEG recording malfunction, leaving $N = 25$ for behavioral analyses; reinforcement-learning model fits used $N = 23$. Artifact-based epoch rejection reduced the neural decoder samples to $N_{\text{REW}} = 24$ and $N_{\text{PUN}} = 25$; a stricter class-balance exclusion for the prevalence analysis in Fig.~2c yielded $N = 22$ (see Extended Data Fig.~9 for sensitivity to this threshold).

Both tasks were two-armed bandits with ${\sim}65$--70\% contingency validity. Correct responses in the reward condition earned $+10$ cents (otherwise 0); incorrect responses in the punishment condition triggered an aversive noise burst. Each task ran for approximately 280 trials. Behavioral commitment was defined as win-stay minus lose-shift probability and tested against zero with a one-sample $t$-test.

EEG preprocessing consisted of bandpass filtering at 1--40~Hz, average referencing, epoching $-200$ to 600~ms surrounding feedback onset, baseline correction, and rejection of epochs exceeding $\pm 150$~$\mu$V (implemented in MNE-Python). Three feedback-locked features entered the decoder: theta power (4--8~Hz at FCz, 200--400~ms), frontal beta power (13--30~Hz, frontal channels, 200--400~ms), and P300 amplitude (Pz, 250--450~ms). Logistic regression decoders (3-fold cross-validation) predicted feedback valence and pre-feedback expectation separately; the neural gap on held-out folds was P(feedback$+$$|$EEG) $-$ P(expectation$+$$|$EEG). Four candidate truth definitions were evaluated -- slow exponential moving average of feedback, pre-feedback expectation ratings, Bayesian HMM-inferred correct side, and fitted RL model Q-values -- and the pre-feedback expectation was selected as the only non-circular truth label decodable above chance in both conditions (Extended Data Fig.~10). Cross-modal associations were assessed with Spearman correlation at the subject level (behavioral commitment vs.\ neural $\AUGpos$) and Pearson correlation on smoothed trial-level time series (15-trial moving average).

\subsection*{Dopaminergic modulation (Extended Data Fig.~7)}

Study 2 of the Cavanagh \& Frank dataset\cite{cavanagh_conflict_2014} included $N = 27$ healthy adults in a double-blind crossover design. Each participant completed the Probabilistic Selection Task under cabergoline (1.25~mg, a D2/D3 agonist) and placebo in separate sessions, with order counterbalanced. Three stimulus pairs offered graded noise regimes (20\%, 30\%, 40\%). The primary measure was the gap regulation slope -- the linear trend of the feedback-minus-truth gap time series (20-trial rolling window) over trials -- quantifying how fast the gap resolves within a session. $\AUGpos$ (cumulative over-commitment) served as a secondary measure. Drug effects were examined with a one-sided paired $t$-test (directional hypothesis: a dopamine agonist should flatten the regulation slope and reduce over-commitment).

\subsection*{Statistical reporting}

All tests are two-tailed unless otherwise noted; the exception is the dopaminergic modulation analysis (Extended Data Fig.~7), where a directional hypothesis was specified before examining the data. Effect sizes are Cohen's $d$ throughout: group mean divided by pooled SD for between-group comparisons, mean difference divided by SD of differences for paired and one-sample tests. Correlations are Spearman $\rho$, each accompanied by 95\% CIs. Normality was assessed with Shapiro--Wilk tests; when distributions deviated, non-parametric alternatives were run alongside the parametric tests (Wilcoxon signed-rank for the drug analysis, Mann--Whitney $U$ for developmental comparisons).

No multiple-comparison correction was applied to the primary confirmatory tests ($\AUGpos > 0$ in each system), because each is a single pre-specified hypothesis tested independently per system. Exploratory correlations (individual differences, cross-modal, developmental) carry exact $p$-values without correction and should be interpreted accordingly. The only covariate in any analysis is the empirical noise rate in the individual-differences partial correlation (Extended Data Fig.~6b). Sample sizes were determined by the publicly available datasets; no power analyses were conducted in advance. ``Truth'' is defined differently across systems: in the ML and PRL analyses, truth is objective (held-out labels and the objectively better option, respectively). In the EEG analysis, the truth reference is the participant's pre-feedback expectation of correctness -- the only non-circular, non-feedback label decodable above chance in both conditions (Extended Data Fig.~10). This is an operational decision and does not imply that the participant's confidence is equivalent to objective truth.

\section*{Data Availability}

All datasets are publicly available: Eckstein \textit{et al.} (2022) PRL data (\url{https://osf.io/7wuh4/}), Stolz \textit{et al.} (2022) EEG data (\url{https://openneuro.org/datasets/ds004295}), Cavanagh \& Frank (2014) PST data (\url{https://openneuro.org/datasets/ds004532}), tabular datasets via UCI/OpenML, CIFAR-10N (\url{https://github.com/UCSC-REAL/cifar-10-100n}). Processed analysis artifacts are provided in the companion repository.

\section*{Code Availability}

Analysis scripts for all results are available at [GitHub repository URL upon acceptance].



\bibliography{references}

@article{frenay_classification_2014,
	title = {Classification in the {Presence} of {Label} {Noise}: {A} {Survey}},
	volume = {25},
	issn = {2162-237X, 2162-2388},
	url = {http://ieeexplore.ieee.org/document/6685834/},
	doi = {10.1109/TNNLS.2013.2292894},
	number = {5},
	journal = {IEEE Transactions on Neural Networks and Learning Systems},
	author = {Frenay, Benoit and Verleysen, Michel},
	month = may,
	year = {2014},
	pages = {845--869},
}

@article{song_learning_2023,
	title = {Learning {From} {Noisy} {Labels} {With} {Deep} {Neural} {Networks}: {A} {Survey}},
	volume = {34},
	issn = {2162-237X, 2162-2388},
	url = {https://ieeexplore.ieee.org/document/9729424/},
	doi = {10.1109/TNNLS.2022.3152527},
	number = {11},
	journal = {IEEE Transactions on Neural Networks and Learning Systems},
	author = {Song, Hwanjun and Kim, Minseok and Park, Dongmin and Shin, Yooju and Lee, Jae-Gil},
	month = nov,
	year = {2023},
	pages = {8135--8153},
}

@article{daw_cortical_2006,
	title = {Cortical substrates for exploratory decisions in humans},
	volume = {441},
	issn = {0028-0836, 1476-4687},
	url = {https://www.nature.com/articles/nature04766},
	doi = {10.1038/nature04766},
	language = {en},
	number = {7095},
	journal = {Nature},
	author = {Daw, Nathaniel D. and O'Doherty, John P. and Dayan, Peter and Seymour, Ben and Dolan, Raymond J.},
	month = jun,
	year = {2006},
	pages = {876--879},
}

@article{hassabis_neuroscience-inspired_2017,
	title = {Neuroscience-{Inspired} {Artificial} {Intelligence}},
	volume = {95},
	issn = {08966273},
	url = {https://linkinghub.elsevier.com/retrieve/pii/S0896627317305093},
	doi = {10.1016/j.neuron.2017.06.011},
	language = {en},
	number = {2},
	journal = {Neuron},
	author = {Hassabis, Demis and Kumaran, Dharshan and Summerfield, Christopher and Botvinick, Matthew},
	month = jul,
	year = {2017},
	pages = {245--258},
}

@article{lake_building_2017,
	title = {Building machines that learn and think like people},
	volume = {40},
	issn = {0140-525X, 1469-1825},
	url = {https://www.cambridge.org/core/product/identifier/S0140525X16001837/type/journal_article},
	doi = {10.1017/S0140525X16001837},
	language = {en},
	journal = {Behavioral and Brain Sciences},
	author = {Lake, Brenden M. and Ullman, Tomer D. and Tenenbaum, Joshua B. and Gershman, Samuel J.},
	year = {2017},
	pages = {e253},
}

@article{richards_deep_2019,
	title = {A deep learning framework for neuroscience},
	volume = {22},
	issn = {1097-6256, 1546-1726},
	url = {https://www.nature.com/articles/s41593-019-0520-2},
	doi = {10.1038/s41593-019-0520-2},
	language = {en},
	number = {11},
	journal = {Nature Neuroscience},
	author = {Richards, Blake A. and Lillicrap, Timothy P. and Beaudoin, Philippe and Bengio, Yoshua and Bogacz, Rafal and Christensen, Amelia and Clopath, Claudia and Costa, Rui Ponte and De Berker, Archy and Ganguli, Surya and Gillon, Colleen J. and Hafner, Danijar and Kepecs, Adam and Kriegeskorte, Nikolaus and Latham, Peter and Lindsay, Grace W. and Miller, Kenneth D. and Naud, Richard and Pack, Christopher C. and Poirazi, Panayiota and Roelfsema, Pieter and Sacramento, João and Saxe, Andrew and Scellier, Benjamin and Schapiro, Anna C. and Senn, Walter and Wayne, Greg and Yamins, Daniel and Zenke, Friedemann and Zylberberg, Joel and Therien, Denis and Kording, Konrad P.},
	month = nov,
	year = {2019},
	pages = {1761--1770},
}

@article{marblestone_toward_2016,
	title = {Toward an {Integration} of {Deep} {Learning} and {Neuroscience}},
	volume = {10},
	issn = {1662-5188},
	url = {http://journal.frontiersin.org/Article/10.3389/fncom.2016.00094/abstract},
	doi = {10.3389/fncom.2016.00094},
	journal = {Frontiers in Computational Neuroscience},
	author = {Marblestone, Adam H. and Wayne, Greg and Kording, Konrad P.},
	month = sep,
	year = {2016},
}

@article{cools_defining_2002,
	title = {Defining the {Neural} {Mechanisms} of {Probabilistic} {Reversal} {Learning} {Using} {Event}-{Related} {Functional} {Magnetic} {Resonance} {Imaging}},
	volume = {22},
	issn = {0270-6474, 1529-2401},
	url = {https://www.jneurosci.org/lookup/doi/10.1523/JNEUROSCI.22-11-04563.2002},
	doi = {10.1523/JNEUROSCI.22-11-04563.2002},
	language = {en},
	number = {11},
	journal = {The Journal of Neuroscience},
	author = {Cools, Roshan and Clark, Luke and Owen, Adrian M. and Robbins, Trevor W.},
	month = jun,
	year = {2002},
	pages = {4563--4567},
}

@article{izquierdo_neural_2017,
	title = {The neural basis of reversal learning: {An} updated perspective},
	volume = {345},
	issn = {03064522},
	url = {https://linkinghub.elsevier.com/retrieve/pii/S030645221600244X},
	doi = {10.1016/j.neuroscience.2016.03.021},
	language = {en},
	journal = {Neuroscience},
	author = {Izquierdo, A. and Brigman, J.L. and Radke, A.K. and Rudebeck, P.H. and Holmes, A.},
	month = mar,
	year = {2017},
	pages = {12--26},
}

@article{schoenbaum_new_2009,
	title = {A new perspective on the role of the orbitofrontal cortex in adaptive behaviour},
	volume = {10},
	issn = {1471-003X, 1471-0048},
	url = {https://www.nature.com/articles/nrn2753},
	doi = {10.1038/nrn2753},
	language = {en},
	number = {12},
	journal = {Nature Reviews Neuroscience},
	author = {Schoenbaum, Geoffrey and Roesch, Matthew R. and Stalnaker, Thomas A. and Takahashi, Yuji K.},
	month = dec,
	year = {2009},
	pages = {885--892},
}

@article{clark_neuropsychology_2004,
	title = {The neuropsychology of ventral prefrontal cortex: {Decision}-making and reversal learning},
	volume = {55},
	issn = {02782626},
	url = {https://linkinghub.elsevier.com/retrieve/pii/S0278262603002847},
	doi = {10.1016/S0278-2626(03)00284-7},
	language = {en},
	number = {1},
	journal = {Brain and Cognition},
	author = {Clark, L. and Cools, R. and Robbins, T.W.},
	month = jun,
	year = {2004},
	pages = {41--53},
}

@article{collins_how_2012,
	title = {How much of reinforcement learning is working memory, not reinforcement learning? {A} behavioral, computational, and neurogenetic analysis},
	volume = {35},
	issn = {0953-816X, 1460-9568},
	url = {https://onlinelibrary.wiley.com/doi/10.1111/j.1460-9568.2011.07980.x},
	doi = {10.1111/j.1460-9568.2011.07980.x},
	language = {en},
	number = {7},
	journal = {European Journal of Neuroscience},
	author = {Collins, Anne G. E. and Frank, Michael J.},
	month = apr,
	year = {2012},
	pages = {1024--1035},
}

@article{niv_neural_2012,
	title = {Neural {Prediction} {Errors} {Reveal} a {Risk}-{Sensitive} {Reinforcement}-{Learning} {Process} in the {Human} {Brain}},
	volume = {32},
	issn = {0270-6474, 1529-2401},
	url = {https://www.jneurosci.org/lookup/doi/10.1523/JNEUROSCI.5498-10.2012},
	doi = {10.1523/JNEUROSCI.5498-10.2012},
	language = {en},
	number = {2},
	journal = {The Journal of Neuroscience},
	author = {Niv, Yael and Edlund, Jeffrey A. and Dayan, Peter and O'Doherty, John P.},
	month = jan,
	year = {2012},
	pages = {551--562},
}

@article{dayan_reward_2002,
	title = {Reward, {Motivation}, and {Reinforcement} {Learning}},
	volume = {36},
	issn = {08966273},
	url = {https://linkinghub.elsevier.com/retrieve/pii/S0896627302009637},
	doi = {10.1016/S0896-6273(02)00963-7},
	language = {en},
	number = {2},
	journal = {Neuron},
	author = {Dayan, Peter and Balleine, Bernard W.},
	month = oct,
	year = {2002},
	pages = {285--298},
}

@article{polich_updating_2007,
	title = {Updating {P300}: {An} integrative theory of {P3a} and {P3b}},
	volume = {118},
	issn = {13882457},
	url = {https://linkinghub.elsevier.com/retrieve/pii/S1388245707001897},
	doi = {10.1016/j.clinph.2007.04.019},
	language = {en},
	number = {10},
	journal = {Clinical Neurophysiology},
	author = {Polich, John},
	month = oct,
	year = {2007},
	pages = {2128--2148},
}

@article{holroyd_neural_2002,
	title = {The neural basis of human error processing: {Reinforcement} learning, dopamine, and the error-related negativity.},
	volume = {109},
	issn = {1939-1471, 0033-295X},
	url = {https://doi.apa.org/doi/10.1037/0033-295X.109.4.679},
	doi = {10.1037/0033-295X.109.4.679},
	language = {en},
	number = {4},
	journal = {Psychological Review},
	author = {Holroyd, Clay B. and Coles, Michael G. H.},
	month = oct,
	year = {2002},
	pages = {679--709},
}

@article{cavanagh_frontal_2014,
	title = {Frontal theta as a mechanism for cognitive control},
	volume = {18},
	issn = {13646613},
	url = {https://linkinghub.elsevier.com/retrieve/pii/S1364661314001077},
	doi = {10.1016/j.tics.2014.04.012},
	language = {en},
	number = {8},
	journal = {Trends in Cognitive Sciences},
	author = {Cavanagh, James F. and Frank, Michael J.},
	month = aug,
	year = {2014},
	pages = {414--421},
}

@article{ridderinkhof_role_2004,
	title = {The {Role} of the {Medial} {Frontal} {Cortex} in {Cognitive} {Control}},
	volume = {306},
	issn = {0036-8075, 1095-9203},
	url = {https://www.science.org/doi/10.1126/science.1100301},
	doi = {10.1126/science.1100301},
	language = {en},
	number = {5695},
	journal = {Science},
	author = {Ridderinkhof, K. Richard and Ullsperger, Markus and Crone, Eveline A. and Nieuwenhuis, Sander},
	month = oct,
	year = {2004},
	pages = {443--447},
}

@article{walsh_learning_2012,
	title = {Learning from experience: {Event}-related potential correlates of reward processing, neural adaptation, and behavioral choice},
	volume = {36},
	issn = {01497634},
	url = {https://linkinghub.elsevier.com/retrieve/pii/S0149763412000875},
	doi = {10.1016/j.neubiorev.2012.05.008},
	language = {en},
	number = {8},
	journal = {Neuroscience \& Biobehavioral Reviews},
	author = {Walsh, Matthew M. and Anderson, John R.},
	month = sep,
	year = {2012},
	pages = {1870--1884},
}

@inproceedings{patrini_making_2017,
	address = {Honolulu, HI},
	title = {Making {Deep} {Neural} {Networks} {Robust} to {Label} {Noise}: {A} {Loss} {Correction} {Approach}},
	isbn = {978-1-5386-0457-1},
	url = {http://ieeexplore.ieee.org/document/8099723/},
	doi = {10.1109/CVPR.2017.240},
	booktitle = {2017 {IEEE} {Conference} on {Computer} {Vision} and {Pattern} {Recognition} ({CVPR})},
	publisher = {IEEE},
	author = {Patrini, Giorgio and Rozza, Alessandro and Menon, Aditya Krishna and Nock, Richard and Qu, Lizhen},
	month = jul,
	year = {2017},
	pages = {2233--2241},
}

@article{gershman_reinforcement_2017,
	title = {Reinforcement {Learning} and {Episodic} {Memory} in {Humans} and {Animals}: {An} {Integrative} {Framework}},
	volume = {68},
	issn = {0066-4308, 1545-2085},
	url = {https://www.annualreviews.org/doi/10.1146/annurev-psych-122414-033625},
	doi = {10.1146/annurev-psych-122414-033625},
	language = {en},
	number = {1},
	journal = {Annual Review of Psychology},
	author = {Gershman, Samuel J. and Daw, Nathaniel D.},
	month = jan,
	year = {2017},
	pages = {101--128},
}

@article{cavanagh_conflict_2014,
	title = {Conflict acts as an implicit cost in reinforcement learning},
	volume = {5},
	issn = {2041-1723},
	url = {https://www.nature.com/articles/ncomms6394},
	doi = {10.1038/ncomms6394},
	language = {en},
	number = {1},
	journal = {Nature Communications},
	author = {Cavanagh, James F. and Masters, Sean E. and Bath, Kevin and Frank, Michael J.},
	month = nov,
	year = {2014},
	pages = {5394},
}

@article{schultz_neural_1997,
	title = {A {Neural} {Substrate} of {Prediction} and {Reward}},
	volume = {275},
	issn = {0036-8075, 1095-9203},
	url = {https://www.science.org/doi/10.1126/science.275.5306.1593},
	doi = {10.1126/science.275.5306.1593},
	language = {en},
	number = {5306},
	journal = {Science},
	author = {Schultz, Wolfram and Dayan, Peter and Montague, P. Read},
	month = mar,
	year = {1997},
	pages = {1593--1599},
}

@article{schultz_dopamine_2016,
	title = {Dopamine reward prediction-error signalling: a two-component response},
	volume = {17},
	issn = {1471-003X, 1471-0048},
	url = {https://www.nature.com/articles/nrn.2015.26},
	doi = {10.1038/nrn.2015.26},
	language = {en},
	number = {3},
	journal = {Nature Reviews Neuroscience},
	author = {Schultz, Wolfram},
	month = mar,
	year = {2016},
	pages = {183--195},
}

@article{frank_by_2004,
	title = {By {Carrot} or by {Stick}: {Cognitive} {Reinforcement} {Learning} in {Parkinsonism}},
	volume = {306},
	issn = {0036-8075, 1095-9203},
	url = {https://www.science.org/doi/10.1126/science.1102941},
	doi = {10.1126/science.1102941},
	language = {en},
	number = {5703},
	journal = {Science},
	author = {Frank, Michael J. and Seeberger, Lauren C. and O'Reilly, Randall C.},
	month = dec,
	year = {2004},
	pages = {1940--1943},
}

@article{cools_dopaminergic_2006,
	title = {Dopaminergic modulation of cognitive function-implications for l-{DOPA} treatment in {Parkinson}'s disease},
	volume = {30},
	issn = {01497634},
	url = {https://linkinghub.elsevier.com/retrieve/pii/S0149763405000540},
	doi = {10.1016/j.neubiorev.2005.03.024},
	language = {en},
	number = {1},
	journal = {Neuroscience \& Biobehavioral Reviews},
	author = {Cools, Roshan},
	month = jan,
	year = {2006},
	pages = {1--23},
}

@article{frank_mechanistic_2006,
	title = {A mechanistic account of striatal dopamine function in human cognition: {Psychopharmacological} studies with cabergoline and haloperidol.},
	volume = {120},
	issn = {1939-0084, 0735-7044},
	url = {https://doi.apa.org/doi/10.1037/0735-7044.120.3.497},
	doi = {10.1037/0735-7044.120.3.497},
	language = {en},
	number = {3},
	journal = {Behavioral Neuroscience},
	author = {Frank, Michael J. and O'Reilly, Randall C.},
	year = {2006},
	pages = {497--517},
}

@article{casey_beyond_2015,
	title = {Beyond {Simple} {Models} of {Self}-{Control} to {Circuit}-{Based} {Accounts} of {Adolescent} {Behavior}},
	volume = {66},
	issn = {0066-4308, 1545-2085},
	url = {https://www.annualreviews.org/doi/10.1146/annurev-psych-010814-015156},
	doi = {10.1146/annurev-psych-010814-015156},
	language = {en},
	number = {1},
	journal = {Annual Review of Psychology},
	author = {Casey, B. J.},
	month = jan,
	year = {2015},
	pages = {295--319},
}

@article{somerville_developmental_2010,
	title = {Developmental neurobiology of cognitive control and motivational systems},
	volume = {20},
	issn = {09594388},
	url = {https://linkinghub.elsevier.com/retrieve/pii/S0959438810000073},
	doi = {10.1016/j.conb.2010.01.006},
	language = {en},
	number = {2},
	journal = {Current Opinion in Neurobiology},
	author = {Somerville, Leah H and Casey, Bj},
	month = apr,
	year = {2010},
	pages = {236--241},
}

@article{hartley_neuroscience_2015,
	title = {The neuroscience of adolescent decision-making},
	volume = {5},
	issn = {23521546},
	url = {https://linkinghub.elsevier.com/retrieve/pii/S2352154615001205},
	doi = {10.1016/j.cobeha.2015.09.004},
	language = {en},
	journal = {Current Opinion in Behavioral Sciences},
	author = {Hartley, Catherine A and Somerville, Leah H},
	month = oct,
	year = {2015},
	pages = {108--115},
}

@article{waltz_probabilistic_2007,
	title = {Probabilistic reversal learning impairments in schizophrenia: {Further} evidence of orbitofrontal dysfunction},
	volume = {93},
	issn = {09209964},
	url = {https://linkinghub.elsevier.com/retrieve/pii/S092099640700120X},
	doi = {10.1016/j.schres.2007.03.010},
	language = {en},
	number = {1-3},
	journal = {Schizophrenia Research},
	author = {Waltz, James A. and Gold, James M.},
	month = jul,
	year = {2007},
	pages = {296--303},
}

@article{peterson_probabilistic_2009,
	title = {Probabilistic reversal learning is impaired in {Parkinson}'s disease},
	volume = {163},
	issn = {03064522},
	url = {https://linkinghub.elsevier.com/retrieve/pii/S0306452209012068},
	doi = {10.1016/j.neuroscience.2009.07.033},
	language = {en},
	number = {4},
	journal = {Neuroscience},
	author = {Peterson, D.A. and Elliott, C. and Song, D.D. and Makeig, S. and Sejnowski, T.J. and Poizner, H.},
	month = nov,
	year = {2009},
	pages = {1092--1101},
}

@article{maia_reinforcement_2011,
	title = {From reinforcement learning models to psychiatric and neurological disorders},
	volume = {14},
	issn = {1097-6256, 1546-1726},
	url = {https://www.nature.com/articles/nn.2723},
	doi = {10.1038/nn.2723},
	language = {en},
	number = {2},
	journal = {Nature Neuroscience},
	author = {Maia, Tiago V and Frank, Michael J},
	month = feb,
	year = {2011},
	pages = {154--162},
}

@article{murray_substantia_2008,
	title = {Substantia nigra/ventral tegmental reward prediction error disruption in psychosis},
	volume = {13},
	issn = {1359-4184, 1476-5578},
	url = {https://www.nature.com/articles/4002058},
	doi = {10.1038/sj.mp.4002058},
	language = {en},
	number = {3},
	journal = {Molecular Psychiatry},
	author = {Murray, G K and Corlett, P R and Clark, L and Pessiglione, M and Blackwell, A D and Honey, G and Jones, P B and Bullmore, E T and Robbins, T W and Fletcher, P C},
	month = mar,
	year = {2008},
	pages = {267--276},
}

@article{waltz_patients_2009,
	title = {Patients with {Schizophrenia} have a {Reduced} {Neural} {Response} to {Both} {Unpredictable} and {Predictable} {Primary} {Reinforcers}},
	volume = {34},
	issn = {0893-133X, 1740-634X},
	url = {https://www.nature.com/articles/npp2008214},
	doi = {10.1038/npp.2008.214},
	language = {en},
	number = {6},
	journal = {Neuropsychopharmacology},
	author = {Waltz, James A and Schweitzer, Julie B and Gold, James M and Kurup, Pradeep K and Ross, Thomas J and Jo Salmeron, Betty and Rose, Emma Jane and McClure, Samuel M and Stein, Elliot A},
	month = may,
	year = {2009},
	pages = {1567--1577},
}

@article{rendell_why_2010,
	title = {Why {Copy} {Others}? {Insights} from the {Social} {Learning} {Strategies} {Tournament}},
	volume = {328},
	issn = {0036-8075, 1095-9203},
	url = {https://www.science.org/doi/10.1126/science.1184719},
	doi = {10.1126/science.1184719},
	language = {en},
	number = {5975},
	journal = {Science},
	author = {Rendell, L. and Boyd, R. and Cownden, D. and Enquist, M. and Eriksson, K. and Feldman, M. W. and Fogarty, L. and Ghirlanda, S. and Lillicrap, T. and Laland, K. N.},
	month = apr,
	year = {2010},
	pages = {208--213},
}

@article{belkin_reconciling_2019,
	title = {Reconciling modern machine-learning practice and the classical bias–variance trade-off},
	volume = {116},
	issn = {0027-8424, 1091-6490},
	url = {https://pnas.org/doi/full/10.1073/pnas.1903070116},
	doi = {10.1073/pnas.1903070116},
	language = {en},
	number = {32},
	journal = {Proceedings of the National Academy of Sciences},
	author = {Belkin, Mikhail and Hsu, Daniel and Ma, Siyuan and Mandal, Soumik},
	month = aug,
	year = {2019},
	pages = {15849--15854},
}

@article{saxe_mathematical_2019,
	title = {A mathematical theory of semantic development in deep neural networks},
	volume = {116},
	issn = {0027-8424, 1091-6490},
	url = {https://pnas.org/doi/full/10.1073/pnas.1820226116},
	doi = {10.1073/pnas.1820226116},
	language = {en},
	number = {23},
	journal = {Proceedings of the National Academy of Sciences},
	author = {Saxe, Andrew M. and McClelland, James L. and Ganguli, Surya},
	month = jun,
	year = {2019},
	pages = {11537--11546},
}

@misc{arpit_closer_2017,
	title = {A {Closer} {Look} at {Memorization} in {Deep} {Networks}},
	url = {http://arxiv.org/abs/1706.05394},
	doi = {10.48550/arXiv.1706.05394},
	publisher = {arXiv},
	author = {Arpit, Devansh and Jastrzębski, Stanisław and Ballas, Nicolas and Krueger, David and Bengio, Emmanuel and Kanwal, Maxinder S. and Maharaj, Tegan and Fischer, Asja and Courville, Aaron and Bengio, Yoshua and Lacoste-Julien, Simon},
	month = jul,
	year = {2017},
	note = {arXiv:1706.05394 [stat]},
}

@misc{zhang_understanding_2017,
	title = {Understanding deep learning requires rethinking generalization},
	url = {http://arxiv.org/abs/1611.03530},
	doi = {10.48550/arXiv.1611.03530},
	publisher = {arXiv},
	author = {Zhang, Chiyuan and Bengio, Samy and Hardt, Moritz and Recht, Benjamin and Vinyals, Oriol},
	month = feb,
	year = {2017},
	note = {arXiv:1611.03530 [cs]},
}

@book{sutton_reinforcement_2020,
	address = {Cambridge, Massachusetts London, England},
	edition = {Second edition},
	series = {Adaptive computation and machine learning},
	title = {Reinforcement learning: an introduction},
	isbn = {978-0-262-03924-6},
	language = {eng},
	publisher = {The MIT Press},
	author = {Sutton, Richard S. and Barto, Andrew},
	year = {2020},
}

@misc{feldman_what_2020,
	title = {What {Neural} {Networks} {Memorize} and {Why}: {Discovering} the {Long} {Tail} via {Influence} {Estimation}},
	url = {http://arxiv.org/abs/2008.03703},
	doi = {10.48550/arXiv.2008.03703},
	publisher = {arXiv},
	author = {Feldman, Vitaly and Zhang, Chiyuan},
	month = aug,
	year = {2020},
	note = {arXiv:2008.03703 [cs]},
}

@misc{toneva_empirical_2019,
	title = {An {Empirical} {Study} of {Example} {Forgetting} during {Deep} {Neural} {Network} {Learning}},
	url = {http://arxiv.org/abs/1812.05159},
	doi = {10.48550/arXiv.1812.05159},
	publisher = {arXiv},
	author = {Toneva, Mariya and Sordoni, Alessandro and Combes, Remi Tachet des and Trischler, Adam and Bengio, Yoshua and Gordon, Geoffrey J.},
	month = nov,
	year = {2019},
	note = {arXiv:1812.05159 [cs]},
}

@article{siegel_widespread_1996,
	title = {The widespread influence of the {Rescorla}-{Wagner} model},
	volume = {3},
	issn = {1069-9384, 1531-5320},
	url = {http://link.springer.com/10.3758/BF03210755},
	doi = {10.3758/BF03210755},
	language = {en},
	number = {3},
	journal = {Psychonomic Bulletin \& Review},
	author = {Siegel, Shepard and Allan, Lorraine G.},
	month = sep,
	year = {1996},
	pages = {314--321},
}

@incollection{rescorla_theory_1972,
	title = {A theory of {Pavlovian} conditioning: {Variations} in the effectiveness of reinforcement and nonreinforcement},
	volume = {Vol. 2},
	booktitle = {Classical {Conditioning} {II}: {Current} {Research} and {Theory}},
	author = {Rescorla, RA and Wagner, Allan},
	month = jan,
	year = {1972},
	note = {Journal Abbreviation: Classical Conditioning II: Current Research and Theory},
}

@misc{stolz_reward_2022,
	title = {Reward gain and punishment avoidance reversal learning},
	url = {https://openneuro.org/datasets/ds004295/versions/1.0.0},
	doi = {10.18112/OPENNEURO.DS004295.V1.0.0},
	publisher = {Openneuro},
	author = {Stolz, Christopher and Pickering, Alan and Mueller, Erik M.},
	year = {2022},
}

@book{richerson_culture_1985,
	address = {Chicago London},
	title = {Culture and the evolutionary process},
	isbn = {978-0-226-06931-9},
	language = {eng},
	publisher = {University of Chicago press},
	author = {Richerson, Peter J. and Boyd, Robert},
	year = {1985},
}

@misc{neyshabur_exploring_2017,
	title = {Exploring {Generalization} in {Deep} {Learning}},
	url = {http://arxiv.org/abs/1706.08947},
	doi = {10.48550/arXiv.1706.08947},
	publisher = {arXiv},
	author = {Neyshabur, Behnam and Bhojanapalli, Srinadh and McAllester, David and Srebro, Nathan},
	month = jul,
	year = {2017},
	note = {arXiv:1706.08947 [cs]},
}

@article{eckstein_reinforcement_2022,
	title = {Reinforcement learning and {Bayesian} inference provide complementary models for the unique advantage of adolescents in stochastic reversal},
	volume = {55},
	issn = {18789293},
	url = {https://linkinghub.elsevier.com/retrieve/pii/S1878929322000494},
	doi = {10.1016/j.dcn.2022.101106},
	language = {en},
	journal = {Developmental Cognitive Neuroscience},
	author = {Eckstein, Maria K. and Master, Sarah L. and Dahl, Ronald E. and Wilbrecht, Linda and Collins, Anne G.E.},
	month = jun,
	year = {2022},
	pages = {101106},
}

@inproceedings{natarajan_learning_2013,
	address = {Red Hook, NY, USA},
	series = {{NIPS}'13},
	title = {Learning with noisy labels},
	booktitle = {Proceedings of the 27th {International} {Conference} on {Neural} {Information} {Processing} {Systems} - {Volume} 1},
	publisher = {Curran Associates Inc.},
	author = {Natarajan, Nagarajan and Dhillon, Inderjit S. and Ravikumar, Pradeep and Tewari, Ambuj},
	year = {2013},
	note = {event-place: Lake Tahoe, Nevada},
	pages = {1196--1204},
}


\setlength{\parindent}{0pt}

\clearpage
\textbf{Figure 1 $|$ The feedback--truth gap appears across machines and humans.}
\begin{center}
\includegraphics[width=0.95\textwidth]{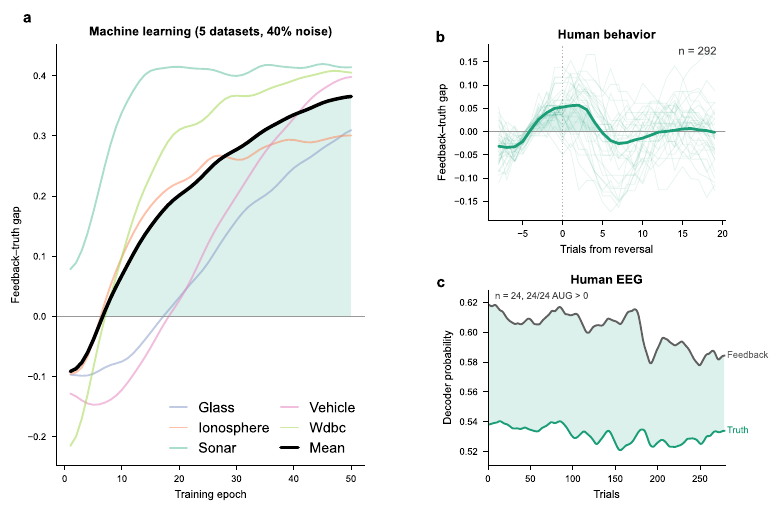}
\end{center}
Each panel plots feedback-aligned minus truth-aligned performance in a system that receives noisy supervision. All three systems produce the gap; only the temporal trajectory differs.
\textbf{a,} An unregularized dense neural network trained on tabular data with 40\% label noise (representative run at the 46th percentile of gap magnitude across $n = 90$ model-noise configurations; Ionosphere dataset; see Extended Data Fig.~1). Training accuracy (grey) exceeds validation accuracy (teal), and the resulting gap (shaded) persists and grows throughout training with no sign of self-correction.
\textbf{b,} Human reversal learning ($N = 292$ subjects, 1,764 reversal epochs from the Eckstein \textit{et al.} dataset; ref.~11). The baseline-corrected feedback--truth gap $\Delta\text{gap}(t)$ is aligned to reversal ($t = 0$). A transient positive deflection (over-commitment) appears immediately post-reversal and returns toward baseline within ${\sim}5$--10 trials (8-trial rolling window; baseline $=$ mean pre-reversal gap).
\textbf{c,} Decoder-based neural gap during reward learning ($n = 24$ subjects from the Stolz \textit{et al.} EEG dataset; ref.~12). Logistic regression decoders trained on theta power, beta power, and P300 amplitude yield P(feedback$+$$|$EEG) (grey) and P(expectation$+$$|$EEG) (teal), where expectation = pre-feedback confidence in the chosen option. The gap between decoder probabilities diminishes across trials (15-trial rolling mean), consistent with progressive neural adaptation; the shading transition marks the onset of gap narrowing. Color convention for all panels: teal $=$ truth, grey $=$ feedback, shaded $=$ learning gap.

\clearpage

\textbf{Figure 2 $|$ The feedback--truth gap is universal across systems and mathematically inevitable.}
\begin{center}
\includegraphics[width=0.95\textwidth]{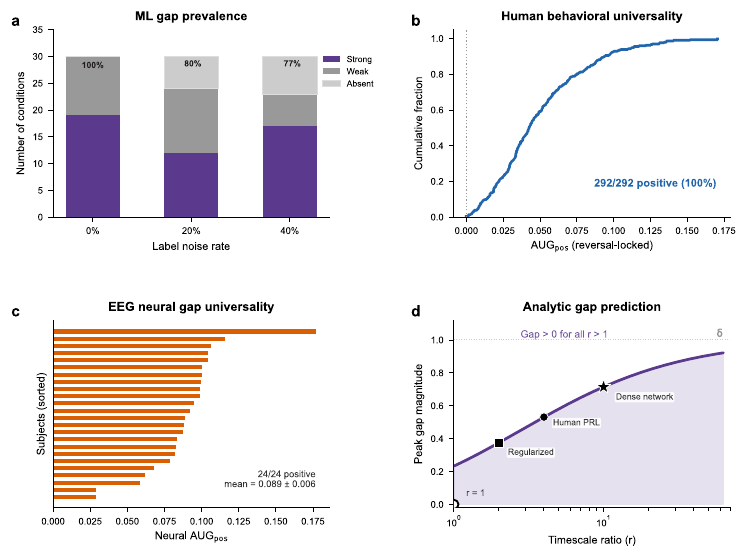}
\end{center}
Gap prevalence across machines and humans, together with the two-timescale model that predicts it.
\textbf{a,} ML prevalence across 90 conditions (5 datasets $\times$ 6 architectures $\times$ 3 noise rates). Stacked bars classify each condition as strong gap (dark purple, AUG $\geq 0.05$), weak (light purple, $0.005 \leq$ AUG $< 0.05$), or absent (grey, AUG $< 0.005$). Across all 90 conditions, 77 (86\%) show a measurable gap. Gap prevalence increases with noise rate and exceeds 90\% at 40\% noise.
\textbf{b,} Cumulative distribution of over-commitment gap (AUG$_{\text{pos}}$) across $N = 292$ PRL subjects. Every subject shows gap $> 0$ (292/292, 100\%). Vertical dashed lines mark quartiles; median AUG$_{\text{pos}} = 0.043$.
\textbf{c,} Per-subject neural gap (AUG$+$ fraction, feedback dominance) in the reward task ($N = 22$ after class-balance exclusion; see Methods and Extended Data Fig.~9), sorted by magnitude. 21/22 subjects (95\%) show feedback-dominant neural representation (mean $\pm$ SEM $= 0.82 \pm 0.03$). Orange bars are individual subjects; white dashed line marks 50\% (parity).
\textbf{d,} Peak gap magnitude from the closed-form solution plotted against timescale ratio $r = \alpha_{\text{fast}} / \alpha_{\text{slow}}$ (log scale). Purple curve: $\text{gap}_{\max} = \delta \cdot r^{-1/(r-1)} \cdot (1 - r^{-1})$. The gap is positive for all $r > 1$ and equals zero exactly at $r = 1$ (matched timescales, black circle). Empirical systems overlaid: Dense network (star), Human PRL (circle), Regularized network (square). See Extended Data Fig.~2c for the full phase diagram.

\clearpage

\textbf{Figure 3 $|$ The gap indexes damage in unregulated networks but structured engagement in regulated humans.}
\begin{center}
\includegraphics[width=0.95\textwidth]{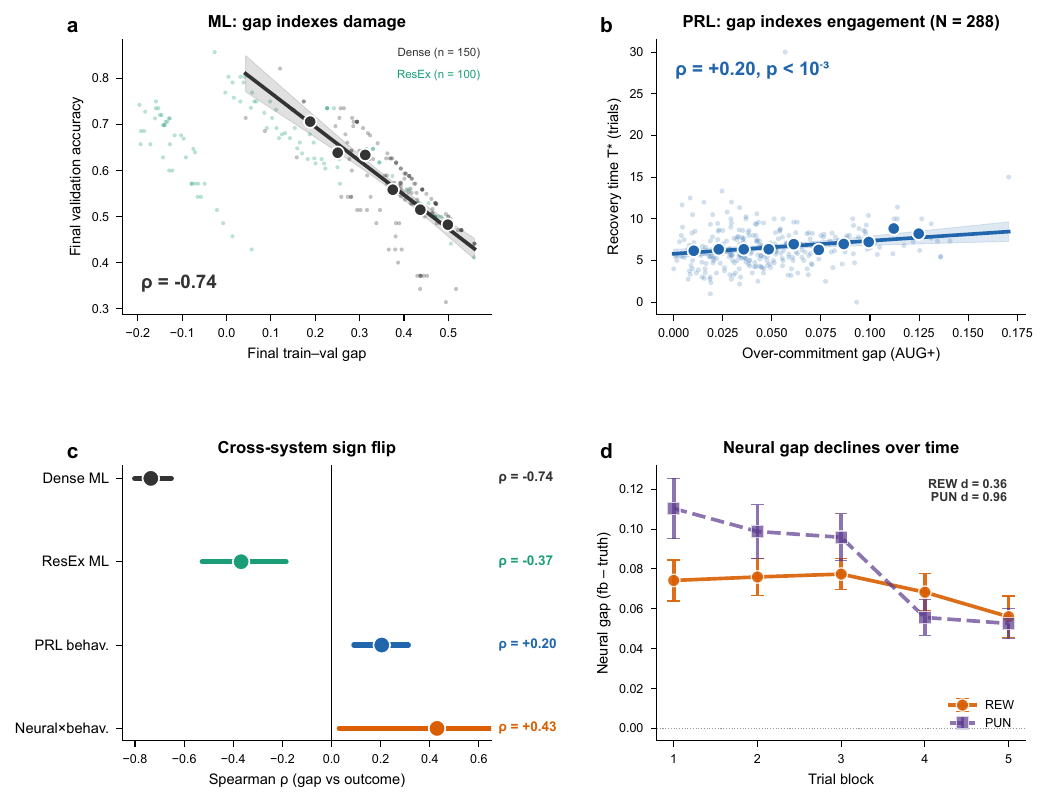}
\end{center}
Whether the gap predicts harm or engagement depends on whether the system regulates it.
\textbf{a,} In unregulated dense networks, a larger final train-validation gap predicts lower validation accuracy (150 dense runs, black; 100 sparse-residual runs, teal). Dense networks: Spearman $\rho = -0.74$. Regression line with 95\% CI shown for dense networks.
\textbf{b,} In human PRL the relationship reverses ($N = 288$ with sufficient post-reversal data). Larger over-commitment (AUG$_{\text{pos}}$) predicts faster recovery ($T^{*}$ in trials; Spearman $\rho = +0.20$, $p < 10^{-3}$), suggesting that the gap reflects contingency engagement rather than damage. Regression line with 95\% CI; point sizes proportional to density.
\textbf{c,} Forest plot of gap--outcome Spearman correlations across four systems. Dense ML: $\rho = -0.74$ (harm); ResEx ML: $\rho = -0.37$ (attenuated); PRL behavioral: $\rho = +0.20$ (sign reversal); Neural $\times$ behavioral: $\rho = +0.43$ (concordant amplification). 95\% CIs shown.
\textbf{d,} Mean neural gap (feedback minus truth decoder probability) across five trial blocks. Reward (orange): $d = 0.36$; punishment (purple): $d = 0.82$. Error bars: $\pm 1$ SEM. The declining trajectory indicates progressive neural recalibration toward truth, distinct from the persistent accumulation seen in networks and the behavioral recovery observed in reversal learning.

\clearpage

\textbf{Figure 4 $|$ The gap is causally tunable via architectural intervention.}
\begin{center}
\includegraphics[width=0.95\textwidth]{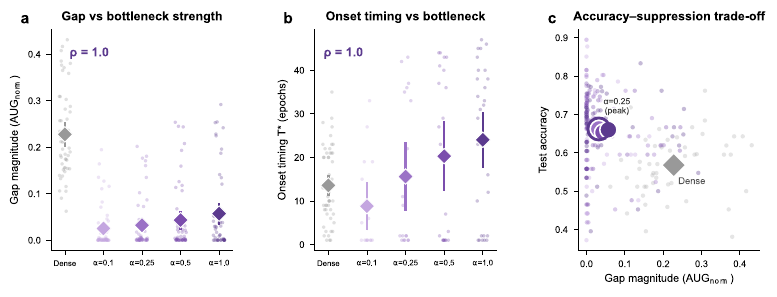}
\end{center}
The sparse-residual bottleneck parameter $\alpha$ provides graded, monotonic control over memorization dynamics and reveals a dissociation between gap suppression and generalization.
\textbf{a,} Normalized cumulative gap (AUG$_{\text{norm}}$) increases monotonically with $\alpha$ (Spearman $\rho = 1.0$ across all datasets with measurable gaps). The dense baseline (grey diamond) shows the largest gap. Light points are individual runs; diamonds are group means with 95\% CI. Five datasets, 40\% label noise, 10 seeds per configuration.
\textbf{b,} Gap onset time ($T^{*}$ in epochs) also increases monotonically with $\alpha$ ($\rho = 1.0$): more constrained architectures delay memorization. Dense baseline (grey diamond) shows earliest onset. Error bars: 95\% CI.
\textbf{c,} Test accuracy plotted against gap magnitude (AUG$_{\text{norm}}$). The relationship is non-monotonic -- accuracy peaks at $\alpha = 0.25$ (teal), not at maximal gap suppression. At $\alpha = 1.0$ (dark teal), accuracy falls below the moderate-suppression level, exposing a capacity--memorization trade-off. Dense baseline (grey diamond) achieves lower accuracy with larger gap. Point sizes proportional to sample count; error bars: 95\% CI.

\clearpage

\textbf{Figure 5 $|$ Neural--behavioral concordance in the feedback--truth gap.}
\begin{center}
\includegraphics[width=0.95\textwidth]{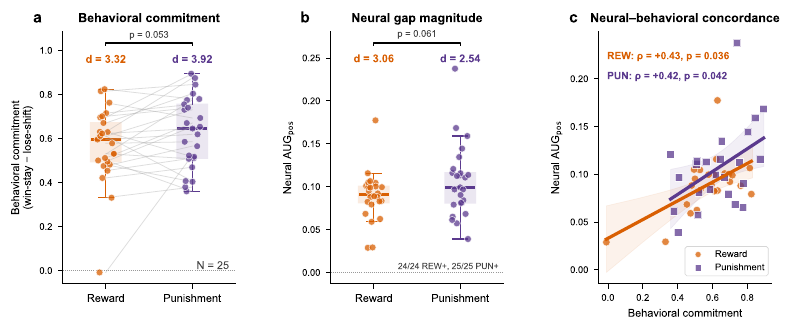}
\end{center}
EEG dataset ($N = 25$; ref.~12). The roughly 10-fold amplification from neural representation to overt behavior is the central observation.
\textbf{a,} Behavioral commitment (win-stay minus lose-shift) for reward (orange, $d = 3.32$) and punishment (purple, $d = 3.92$). Every subject shows positive commitment in both conditions. Connected lines are paired within-subject values; box plots show median and IQR. Difference between tasks: $d = 0.40$, $p = 0.053$ (two-sided paired $t$-test).
\textbf{b,} Neural gap magnitude (AUG$_{\text{pos}}$ from EEG decoders) is roughly an order of magnitude smaller than the behavioral effect. Reward: $d = 1.03$ ($N = 24$); punishment: $d = 1.82$ ($N = 25$). All subjects positive in both conditions (24/24 reward, 25/25 punishment). Box plots with individual data; error bars: $\pm 1$ SEM. Difference: $d = 0.39$, $p = 0.061$.
\textbf{c,} Subjects with larger neural over-representation also show stronger behavioral commitment. Reward: Spearman $\rho = +0.43$, $p = 0.036$ ($n = 24$); punishment: $\rho = +0.42$, $p = 0.042$ ($n = 24$). Regression lines with 95\% CI shown per task. Orange circles: reward; purple squares: punishment.

\clearpage

\textbf{Extended Data Figure 1 $|$ Gap emergence rates across systems.}
\begin{center}
\includegraphics[width=0.95\textwidth]{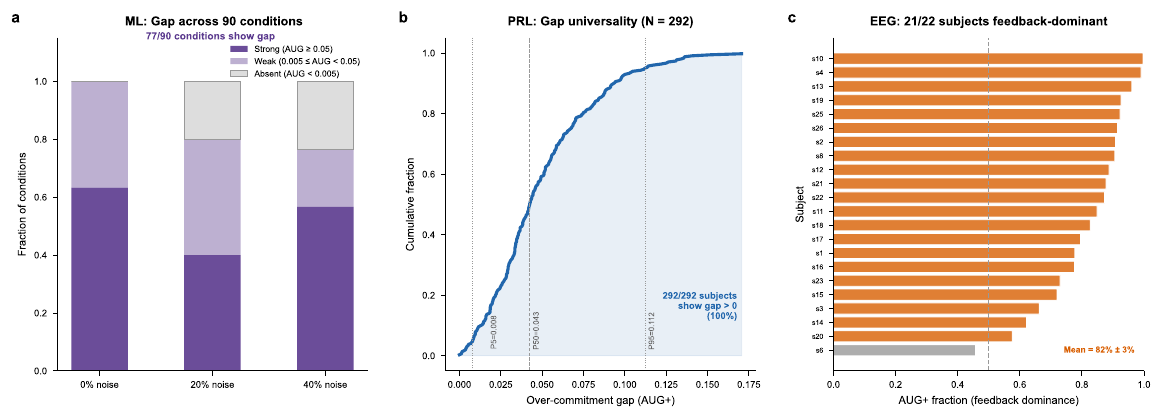}
\end{center}
Under noisy conditions the gap is near-universal. Gap $=$ feedback-aligned minus truth-aligned performance throughout.
\textbf{a,} Fraction of ML conditions showing gap (AUG $\geq 0.005$) at each noise rate (5 datasets $\times$ 6 architectures $\times$ 3 noise rates $= 90$ conditions). Across all 90 conditions, 77 (86\%) show a measurable gap; the proportion exceeds 90\% at 40\% noise. Bars are proportions; no error bars (each condition is a single count).
\textbf{b,} Cumulative distribution of $\AUGpos$ across $N = 292$ PRL subjects (Eckstein dataset; ref.~11). Virtually all subjects show positive over-commitment (median $\AUGpos = 0.043$). Feedback $=$ reward received; truth $=$ objectively better option.
\textbf{c,} Per-subject AUG$+$ fraction (feedback dominance; $N = 22$ non-excluded subjects from Stolz EEG dataset; ref.~12). 95\% of subjects exceed the 50\% parity line (mean $\pm$ SEM $= 0.82 \pm 0.03$). Feedback $=$ reinforcer valence; truth $=$ pre-feedback expectation.

\clearpage

\textbf{Extended Data Figure 2 $|$ Boundary conditions: when the gap vanishes.}
\begin{center}
\includegraphics[width=0.95\textwidth]{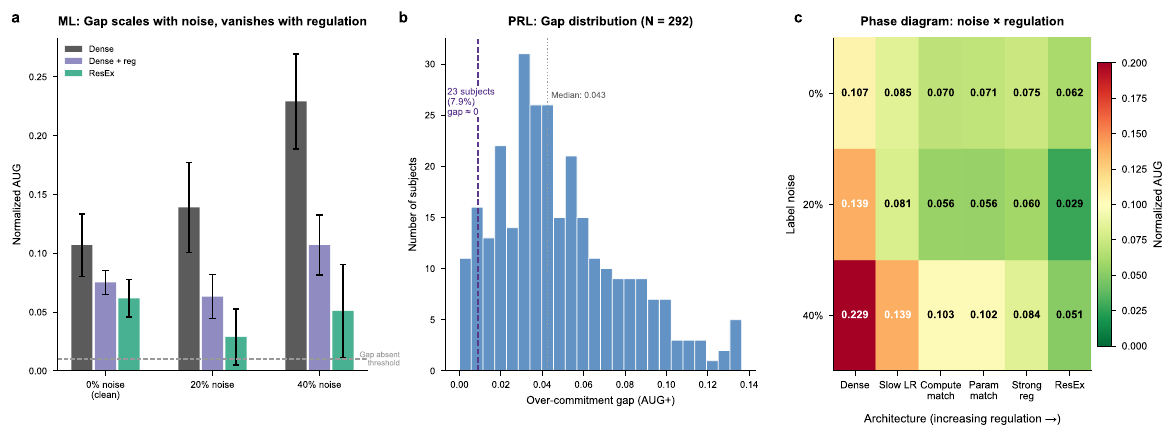}
\end{center}
The inevitability claim is falsifiable: the gap disappears under specific counterfactual conditions. Gap $=$ feedback-aligned minus truth-aligned performance throughout.
\textbf{a,} AUG by noise rate and architecture across 90 ML conditions (5 datasets $\times$ 6 architectures $\times$ 3 noise rates). The gap scales with noise and vanishes under strong regularization or 0\% noise; 15/90 conditions show near-zero gap (AUG $< 0.01$). Points are individual conditions.
\textbf{b,} Distribution of $\AUGpos$ across $N = 292$ PRL subjects. 23 subjects (7.9\%) show near-zero gap, consistent with low-noise or matched-timescale subgroups. Feedback $=$ reward received; truth $=$ objectively better option.
\textbf{c,} Phase diagram showing AUG across noise rate $\times$ architecture. Color scale is mean normalized AUG; cells with white squares have AUG $< 0.01$ (gap absent). The gap vanishes under clean labels, strong regularization, or capacity suppression.

\clearpage

\textbf{Extended Data Figure 3 $|$ Neural validation details.}
\begin{center}
\includegraphics[width=0.95\textwidth]{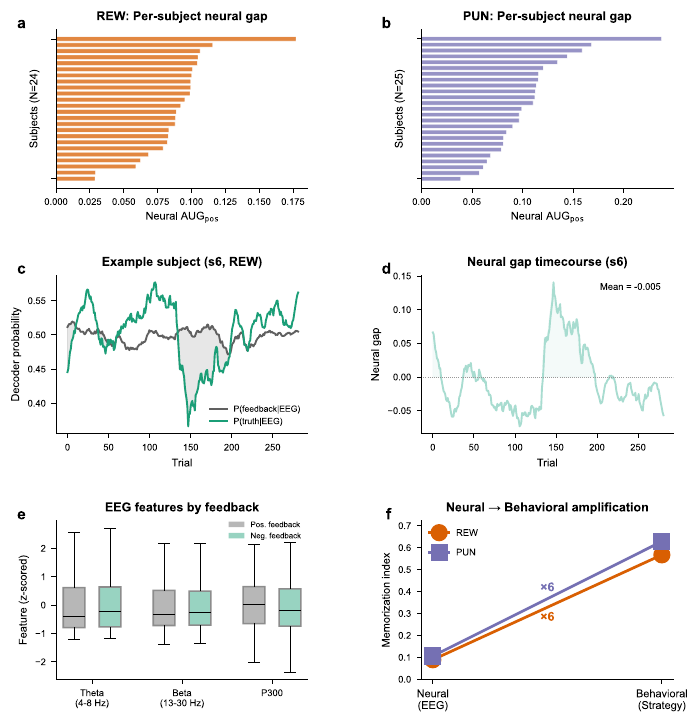}
\end{center}
Subject-level neural gap data, decoder validation, and cross-modal dynamics. Feedback $=$ reinforcer valence (reward or punishment); truth $=$ prior expectation (pre-feedback confidence; see Methods).
\textbf{a--b,} Individual neural $\AUGpos$ values for reward ($N = 24$) and punishment ($N = 25$), sorted by magnitude. All values are positive. Bars are individual subjects.
\textbf{c--d,} An example subject (s6, reward task) illustrates decoder probability trajectories for feedback and expectation (15-trial rolling mean). Shading marks the gap between decoder curves.
\textbf{e,} Distributions of the three EEG features by feedback type: theta power, beta power, and P300 amplitude. Box plots show median and IQR; whiskers extend to $1.5 \times$ IQR.
\textbf{f,} The memorization index is approximately 10-fold larger at the behavioral than neural level (in standardized effect size), illustrating the amplification from cortical representation to overt strategy. Error bars: $\pm 1$ SEM.

\clearpage

\textbf{Extended Data Figure 4 $|$ Early training dynamics predict later memorization regime.}
\begin{center}
\includegraphics[width=0.95\textwidth]{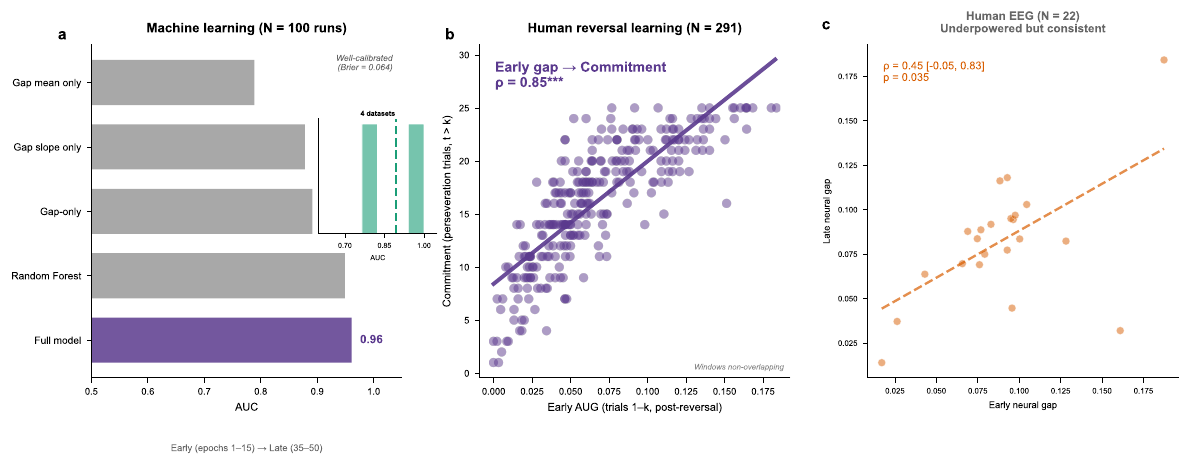}
\end{center}
Features measured early in training (epochs 1--15) predict the later memorization regime (epochs 35+) with a strict 20-epoch temporal buffer, indicating that gap dynamics are deterministic rather than stochastic.
\textbf{a,} Cross-validated AUC across $N = 100$ ML runs (5 datasets, 40\% noise). Logistic regression on 14 early features achieves AUC $= 0.96$ (Brier score $= 0.064$); the single best feature (gap\_slope) alone reaches AUC $= 0.88$. Bars are AUC; error metric is cross-validation standard error. Inset: generalization across datasets.
\textbf{b,} In PRL ($N = 291$; one subject excluded for insufficient post-reversal data), early post-reversal gap predicts later perseveration ($\rho = 0.85$, $p < 0.001$) but not recovery time $\Tstar$ ($\rho = -0.04$, n.s.). Early pressure and downstream regulation are therefore separable.
\textbf{c,} In the EEG sample ($N = 22$), early neural gap predicts later signal persistence ($\rho = 0.45$, $p = 0.035$). The sample is small, but the effect is directionally consistent with the behavioral and ML results.

\clearpage

\textbf{Extended Data Figure 5 $|$ Breadth validation across 30 OpenML datasets.}
\begin{center}
\includegraphics[width=0.95\textwidth]{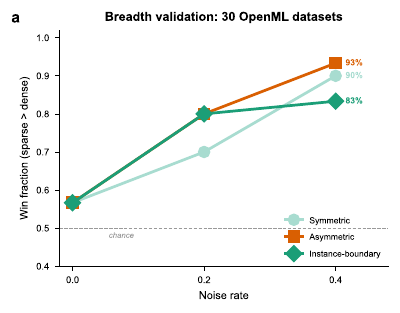}
\end{center}
How often the sparse bottleneck outperforms the dense baseline, plotted against noise rate across 30 OpenML datasets ($n = 2{,}700$ total runs; 30 datasets $\times$ 3 noise rates $\times$ 10 seeds $\times$ 3 configurations). Truth $=$ held-out test labels (uncorrupted); feedback $=$ noisy training labels. At 0\% noise: 57\% (no reliable advantage). At 40\% noise the win fraction rises to 93\% (median $\Delta = +3.28$~pp). Error bars: 95\% binomial CI. The advantage is noise-dependent, not architecture-dominant.

\clearpage

\textbf{Extended Data Figure 6 $|$ Individual differences in gap dynamics.}
\begin{center}
\includegraphics[width=0.95\textwidth]{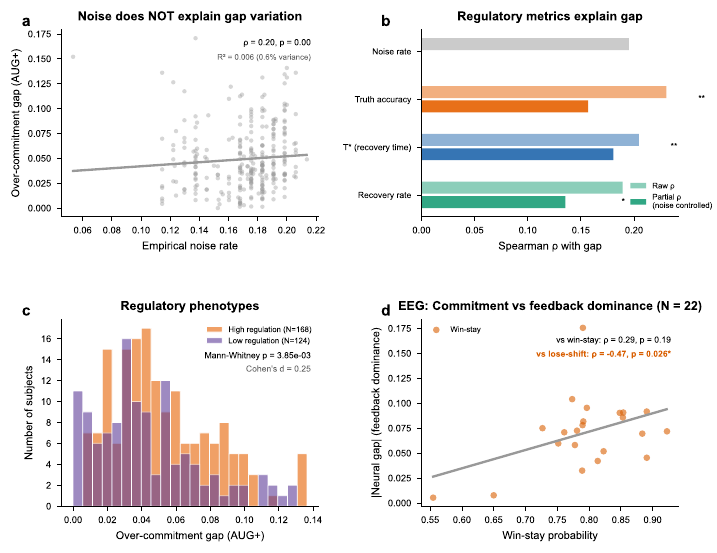}
\end{center}
Individual variation in gap magnitude and recovery speed reveals heterogeneity in regulation. Gap $=$ feedback-aligned minus truth-aligned performance; feedback $=$ reward received; truth $=$ objectively better option (PRL) or pre-feedback expectation (EEG).
\textbf{a,} In PRL ($N = 292$), noise rate positively predicts gap magnitude ($\rho = 0.20$, $p = 0.001$) but explains only 0.6\% of variance -- most of the variation lies between individuals, not between task conditions. Each point is one subject; line is a linear fit.
\textbf{b,} Gap magnitude predicts recovery speed (partial $\rho = 0.14$, $p = 0.021$, controlling for noise rate) and recovery time $\Tstar$ (partial $\rho = 0.18$, $p = 0.002$). PRL, $N = 292$. Each point is one subject.
\textbf{c,} Subjects classified as high-regulation versus low-regulation phenotypes show separable gap profiles ($d = 0.25$). Shading: $\pm 1$ SEM.
\textbf{d,} In the EEG sample ($N = 22$), neural $\AUGpos$ correlates negatively with lose-shift rate ($\rho = -0.48$, $p = 0.026$): subjects with stronger neural over-representation are less likely to shift after losses. Each point is one subject; line is a linear fit.

\clearpage

\textbf{Extended Data Figure 7 $|$ Dopaminergic modulation of the feedback--truth gap.}
\begin{center}
\includegraphics[width=0.95\textwidth]{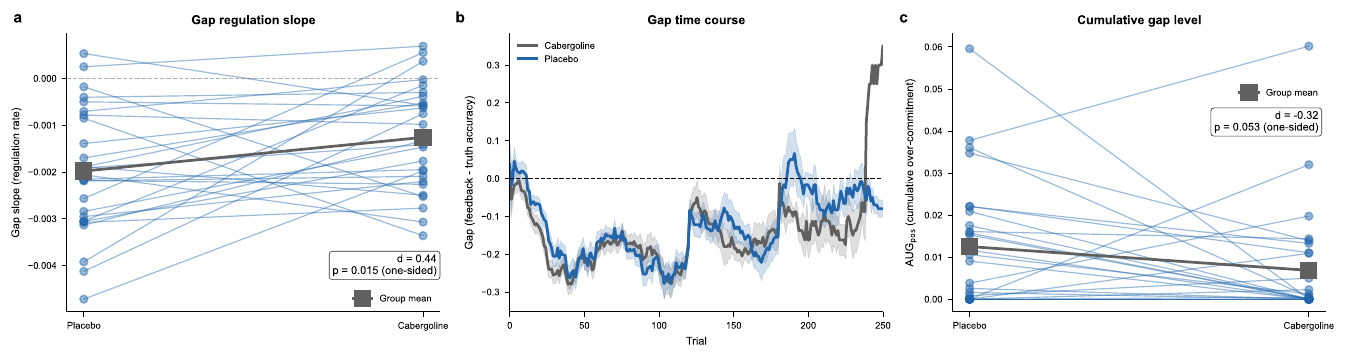}
\end{center}
Cabergoline, a dopamine D2/D3 agonist, flattens within-session gap regulation in a double-blind crossover ($N = 27$; Cavanagh \& Frank dataset; \cite{cavanagh_conflict_2014}). Gap $=$ feedback-aligned minus truth-aligned accuracy (20-trial rolling window).
\textbf{a,} The gap regulation slope (linear trend of gap over trials) is significantly flatter under cabergoline than placebo (one-sided paired $t$-test, $d = 0.44$, $p = 0.015$; Wilcoxon $p = 0.016$). Connected dots are individual subjects; red squares are group means.
\textbf{b,} Group-averaged gap time courses under cabergoline (red) and placebo (blue), with SEM shading. Under placebo the gap declines more steeply (stronger regulation); cabergoline blunts this trajectory.
\textbf{c,} Cumulative over-commitment ($\AUGpos$) is directionally lower under cabergoline ($d = -0.32$, one-sided $p = 0.053$), suggesting that the drug effect extends to both the dynamics and the overall level of the gap.

\clearpage

\textbf{Extended Data Figure 8 $|$ Developmental gradient in feedback persistence and gap regulation.}
\begin{center}
\includegraphics[width=0.95\textwidth]{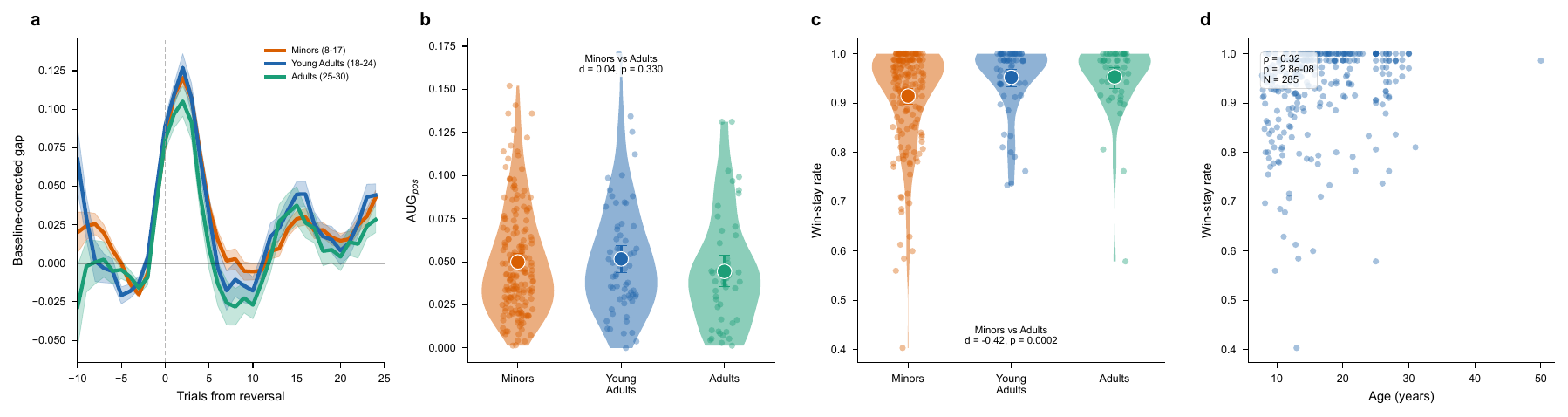}
\end{center}
With development, the balance between flexibility and persistence shifts in a direction that parallels the pharmacological effect of dopamine agonism. Gap $=$ feedback-aligned minus truth-aligned accuracy; feedback $=$ reward received; truth $=$ objectively better option.
\textbf{a,} Reversal-locked gap dynamics for two developmental cohorts (sID $< 300$: minors ages 8--17, $N = 178$; sID $\geq 300$: adults ages 18--30, $N = 114$; Eckstein \textit{et al.} dataset; ref.~11). Mean baseline-corrected $\Delta\text{gap}(t)$, aligned to reversal ($t = 0$). Both groups show a positive post-reversal gap; adults show a slightly flatter recovery trajectory. The faster gap closure in minors reflects reduced feedback persistence rather than enhanced compensatory control. Shading is SEM.
\textbf{b,} Total over-commitment ($\AUGpos$) does not differ significantly between minors and adults ($d = 0.04$, $p = 0.66$, two-sided Mann--Whitney $U$), so the different dynamics in \textbf{a} do not translate into a difference in cumulative magnitude. Violin plots with individual data points; error bars are 95\% bootstrap CI.
\textbf{c,} Win-stay rate (probability of repeating a rewarded choice) increases from minors to adults ($d = -0.42$, $p < 0.001$). The direction parallels the cabergoline effect (Extended Data Fig.~7): both developmental maturation and acute dopamine agonism increase feedback persistence.
\textbf{d,} Treating age as a continuous variable confirms the pattern (Spearman $\rho = 0.32$, $p < 0.0001$; age range 7--50 years, $N = 285$ with matched demographics). The grouped result in \textbf{c} is not an artifact of binning.

\clearpage

\textbf{Extended Data Figure 9 $|$ Cross-modal correlation robustness to exclusion threshold.}
\begin{center}
\includegraphics[width=0.95\textwidth]{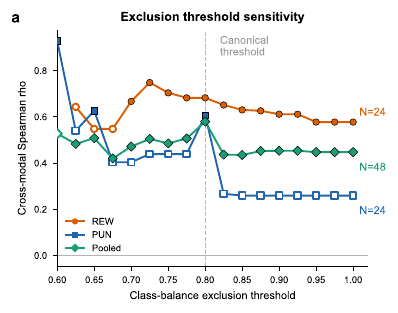}
\end{center}
The cross-modal Spearman correlation (behavioral commitment versus neural $\AUGpos$) is stable across class-balance exclusion thresholds. Subjects whose expectation label imbalance exceeds the threshold are excluded at each step; thresholds from 0.60 to 1.00 were tested. The pooled correlation (reward $+$ punishment) ranges from $\rho = 0.51$ to $\rho = 0.59$ and remains significant ($p < 0.01$) throughout. The 0.80 threshold used in the main analysis falls in the middle of this stable range. Per-condition correlations (reward, punishment) are shown separately. $N$ per threshold varies from 14 to 38 subjects.

\clearpage

\textbf{Extended Data Figure 10 $|$ Non-circular truth reconstruction and decoder comparison.}
\begin{center}
\includegraphics[width=0.95\textwidth]{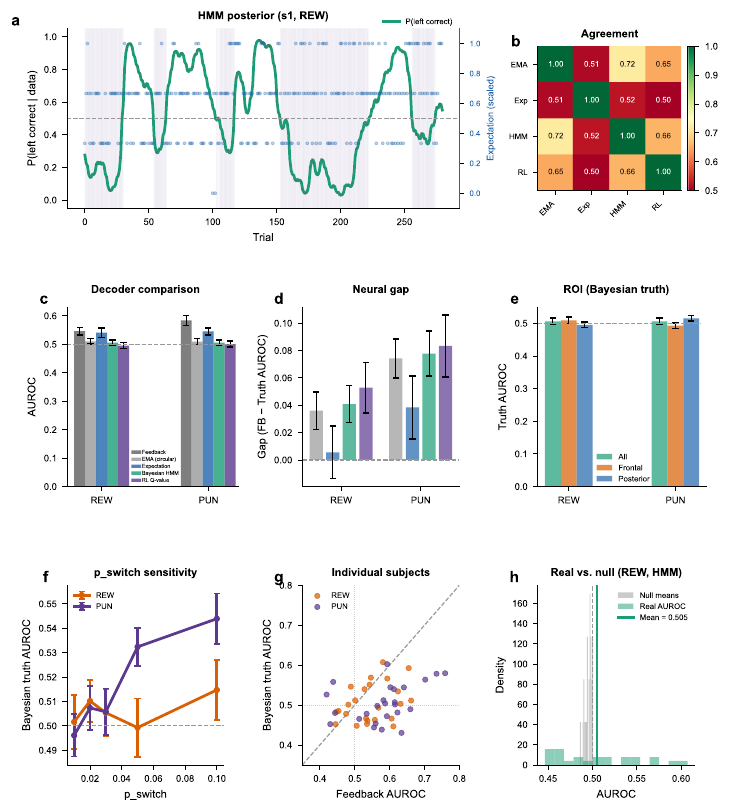}
\end{center}
Post-feedback EEG features decode feedback valence and prior expectation (pre-feedback confidence) but not objective environmental correctness inferred from Bayesian or reinforcement-learning models. We tested four truth definitions: slow exponential moving average of feedback (circular baseline), pre-feedback expectation ratings, Bayesian HMM-inferred correct side (forward--backward with 70/30 emission structure), and fitted RL model $Q$-values (M5 parameters from ref.~12).
\textbf{a,} HMM posterior trajectory for an example subject (s1, REW task): P(left correct $|$ all data), with overlaid expectation ratings (scaled). Purple shading marks epochs where the HMM infers right-correct state.
\textbf{b,} Agreement between truth definitions. Expectation truth is nearly independent of the other three (${\sim}50\%$ agreement), whereas EMA, HMM, and RL cluster together (${\sim}65$--72\%). This is consistent with expectation ratings capturing a genuinely distinct signal.
\textbf{c,} Decoder AUROC across truth definitions and tasks. Only the expectation-truth decoder significantly exceeds chance (REW: AUROC $= 0.541$, $t(23) = 2.43$, $p = 0.012$; PUN: AUROC $= 0.545$, $t(24) = 3.84$, $p < 10^{-3}$). The feedback decoder is shown for reference.
\textbf{d,} Neural gap (feedback minus truth AUROC) by truth definition. With non-circular expectation truth the gap is near-zero in REW (gap $= +0.006$, $p = 0.39$) and small in PUN (gap $= +0.038$, $p = 0.054$), indicating that feedback and expectation representations are nearly balanced in post-feedback EEG.
\textbf{e,} ROI decomposition for Bayesian truth (all features, frontal-only, posterior-only). No feature subset rescues objective-truth decoding.
\textbf{f,} Bayesian truth decoder AUROC across $p_{\text{switch}}$ values (0.01--0.10). The null result for objective truth is stable.
\textbf{g,} Individual-subject scatter of feedback AUROC versus Bayesian truth AUROC. The null is not driven by outlier subjects.
\textbf{h,} Real versus null AUROC distributions for the Bayesian truth decoder (REW task). HMM validation: reward rate when HMM-inferred correct $= 0.86$, when incorrect $= 0.28$, confirming reconstruction validity against the known 70/30 contingency structure.


\clearpage

\textbf{Extended Data Table 1 $|$ Gap dynamics across noise-robust training methods.}

\vspace{6pt}
\begin{tabular}{lcccc}
\hline
Method & Median $\AUGnorm$ & Median $\Tstar$ & Frac.\ AUG $> 0.005$ & Median val.\ acc. \\
\hline
Dense (baseline) & 0.121 & 24.0 & 73\% & 0.646 \\
Sparse-residual & 0.000 & 50.0 (censored) & 30\% & 0.770 \\
Co-teaching & 0.000 & 50.0 (censored) & 23\% & 0.806 \\
Forward loss correction & 0.009 & 48.8 & 57\% & 0.705 \\
\hline
\end{tabular}
\vspace{4pt}

Four training methods applied to 30 OpenML datasets with 40\% symmetric label noise (3 seeds each). Median $\AUGnorm$ and $\Tstar$ computed per dataset (mean across seeds), then median across datasets. All methods produce a measurable gap in a substantial fraction of datasets; no method eliminates it entirely. The gap is suppressed but not abolished by algorithmic intervention, consistent with inevitability under the two-timescale framework.

\clearpage

\textbf{Extended Data Table 2 $|$ Sparse-residual gap suppression across 30 datasets.}

\vspace{6pt}
\resizebox{\textwidth}{!}{%
\begin{tabular}{ccccccc}
\hline
Noise rate & $N$ datasets & Median $\Delta$AUG & Frac.\ $\Delta$AUG $< 0$ & Median $\Delta_{\text{acc}}$ & Frac.\ $\Delta_{\text{acc}} > 0$ & $\Tstar$ delay frac. \\
\hline
0.2 & 30 & $-0.055$ & 80\% [63--93\%] & $+0.052$ & 87\% [73--97\%] & 80\% \\
0.4 & 30 & $-0.119$ & 97\% [90--100\%] & $+0.081$ & 93\% [83--100\%] & 91\% \\
\hline
\end{tabular}}
\vspace{4pt}

Dense baseline versus sparse-residual architecture (hidden dimensions $= 256$, degree $= 5$, $\alpha = 0.25$) across 30 OpenML datasets at two noise rates (3 noise types $\times$ 10 seeds $= 18$ runs per dataset per architecture). $\Delta$AUG $=$ sparse-residual minus dense (negative $=$ suppression). $\Delta_{\text{acc}}$ $=$ sparse-residual minus dense (positive $=$ improvement). $\Tstar$ delay fraction: proportion of datasets where sparse-residual delays the onset of memorization. Sparse-residual models use approximately 5\% of dense-baseline compute (median MACs ratio $= 0.052$).

\clearpage

\textbf{Extended Data Table 3 $|$ Topology null study.}

\vspace{6pt}
\begin{tabular}{lc}
\hline
Metric & Value \\
\hline
$N$ datasets & 10 \\
Median $\Delta\AUGnorm$ (random $-$ expander) & 0.000 \\
Frac.\ $|\Delta\AUGnorm| < 0.01$ & 100\% \\
Median $\Delta_{\text{test acc}}$ & $-0.0004$ \\
Frac.\ ordering consistent (random $\geq$ expander) & 0\% \\
\hline
\end{tabular}
\vspace{4pt}

Expander-graph versus random-regular graph wiring within the same sparse-residual scaffold (10 datasets, symmetric noise at 40\%, 3 seeds). $\Delta\AUGnorm$ $=$ random-regular minus expander. Graph topology produces no detectable difference in gap dynamics: median $\Delta\AUGnorm = 0.000$ and 100\% of datasets fall within $|\Delta| < 0.01$. The sparse-residual scaffolding, not specific graph structure, is the operative ingredient for gap suppression.

\clearpage

\textbf{Extended Data Table 4 $|$ Reinforcement-learning parameters selectively predict gap dynamics.}

\vspace{6pt}
\resizebox{\textwidth}{!}{%
\begin{tabular}{llcccc}
\hline
Outcome & Predictor & $\rho$ & $p$ & Residualized $\rho$ & Residualized $p$ \\
\hline
$\AUGpos$ & $\alpha$ (standard RW) & 0.291 & $< 10^{-6}$ & 0.057 & 0.328 \\
$\AUGpos$ & $\pi$ (perseveration) & $-0.264$ & $< 10^{-5}$ & $-0.083$ & 0.159 \\
$\AUGpos$ & $\alpha^{-}$ (asymmetric) & 0.291 & $< 10^{-6}$ & -- & -- \\
$\AUGpos$ & $\alpha^{+} - \alpha^{-}$ (asymmetry) & $-0.222$ & $< 10^{-4}$ & -- & -- \\
Gap slope & $\alpha$ (standard RW) & 0.333 & $< 10^{-8}$ & 0.367 & $< 10^{-10}$ \\
Gap slope & $\pi$ (perseveration) & $-0.149$ & 0.011 & $-0.207$ & $< 10^{-3}$ \\
Gap slope & $\alpha^{-}$ (asymmetric) & 0.314 & $< 10^{-7}$ & -- & -- \\
Gap slope & $\alpha^{+} - \alpha^{-}$ (asymmetry) & $-0.129$ & 0.027 & $-0.113$ & 0.054 \\
\hline
\end{tabular}}
\vspace{4pt}

Spearman correlations between fitted RL parameters and gap metrics ($N = 292$ subjects, Eckstein 2022 dataset). Three models fitted: standard Rescorla--Wagner (RW), RW with perseveration ($\pi$), and asymmetric RW ($\alpha^{+}$, $\alpha^{-}$). The standard learning rate $\alpha$ predicts both gap magnitude ($\AUGpos$) and recovery speed (gap slope), consistent with the two-timescale framework. Perseveration $\pi$ shows inverse relationships. These associations survive residualization for task-level confounds (noise rate, trial count, reversal count).

\end{document}


\maketitle

\tableofcontents
\clearpage

\section{Supplementary Derivation: Two-Timescale Gap}
\label{sec:derivation}

Below we work out the gap in closed form for the two-timescale learner, show that it is strictly positive whenever the timescales differ, and derive the peak magnitude.

\subsection{Model definition}

A learner tracks an environment state $s \in \{0,1\}$ via two channels, each updated by an exponential moving average:
\begin{align}
x_{\text{fast}}(t+1) &= x_{\text{fast}}(t) + \alpha_{\text{fast}} \cdot \bigl(o(t) - x_{\text{fast}}(t)\bigr) \label{eq:fast} \\
x_{\text{slow}}(t+1) &= x_{\text{slow}}(t) + \alpha_{\text{slow}} \cdot \bigl(s(t) - x_{\text{slow}}(t)\bigr) \label{eq:slow}
\end{align}
where $o(t)$ is the observed feedback signal (equal to $s(t)$ with probability $1 - \varepsilon$ and $1 - s(t)$ with probability $\varepsilon$) and $0 < \alpha_{\text{slow}} < \alpha_{\text{fast}} \leq 1$. In the continuous-time limit, these become exponential decay processes with rates $\alpha_{\text{fast}}$ and $\alpha_{\text{slow}}$ respectively.

\subsection{Gap dynamics after a state change}

Consider a state reversal of magnitude $\Delta$ at time $t = 0$. Both channels begin with the same initial displacement $\Delta$ from the new true state. In expectation, the residual error of each channel decays exponentially:
\begin{align}
\mathbb{E}[\text{error}_{\text{fast}}(t)] &= \Delta \cdot e^{-\alpha_{\text{fast}} \cdot t} \\
\mathbb{E}[\text{error}_{\text{slow}}(t)] &= \Delta \cdot e^{-\alpha_{\text{slow}} \cdot t}
\end{align}

The feedback--truth gap is the difference in alignment between the two channels (equivalently, the difference in residual errors):
\begin{equation}
\text{gap}(t) = \Delta \cdot \bigl(e^{-\alpha_{\text{slow}} \cdot t} - e^{-\alpha_{\text{fast}} \cdot t}\bigr)
\label{eq:gap}
\end{equation}

\subsection{Proof of strict positivity}

\textbf{Proposition 1.} \textit{If $\alpha_{\text{fast}} > \alpha_{\text{slow}} > 0$, then $\text{gap}(t) > 0$ for all $t > 0$.}

\textit{Proof.} Since $\alpha_{\text{fast}} > \alpha_{\text{slow}}$, we have $\alpha_{\text{fast}} \cdot t > \alpha_{\text{slow}} \cdot t$ for all $t > 0$. Because $f(x) = e^{-x}$ is strictly decreasing, $e^{-\alpha_{\text{fast}} \cdot t} < e^{-\alpha_{\text{slow}} \cdot t}$. Therefore $e^{-\alpha_{\text{slow}} \cdot t} - e^{-\alpha_{\text{fast}} \cdot t} > 0$, and since $\Delta > 0$, $\text{gap}(t) > 0$. \hfill $\square$

\textbf{Corollary (Counterfactual).} When $\alpha_{\text{fast}} = \alpha_{\text{slow}}$, $\text{gap}(t) = 0$ for all $t$, regardless of noise level $\varepsilon$.

\subsection{Peak timing}

Differentiating Eq.~\eqref{eq:gap} with respect to $t$ and setting the derivative to zero:
\begin{equation}
\frac{d}{dt}\,\text{gap}(t) = \Delta \cdot \bigl(-\alpha_{\text{slow}} \cdot e^{-\alpha_{\text{slow}} \cdot t} + \alpha_{\text{fast}} \cdot e^{-\alpha_{\text{fast}} \cdot t}\bigr) = 0
\end{equation}

This yields:
\begin{equation}
\alpha_{\text{fast}} \cdot e^{-\alpha_{\text{fast}} \cdot t^*} = \alpha_{\text{slow}} \cdot e^{-\alpha_{\text{slow}} \cdot t^*}
\end{equation}

Taking logarithms:
\begin{equation}
\ln\!\left(\frac{\alpha_{\text{fast}}}{\alpha_{\text{slow}}}\right) = (\alpha_{\text{fast}} - \alpha_{\text{slow}}) \cdot t^*
\end{equation}

Defining the timescale ratio $r = \alpha_{\text{fast}} / \alpha_{\text{slow}}$:
\begin{equation}
t^* = \frac{\ln r}{\alpha_{\text{slow}} \cdot (r - 1)}
\label{eq:tstar}
\end{equation}

\subsection{Peak magnitude}

Substituting $t^*$ back into Eq.~\eqref{eq:gap}:
\begin{align}
e^{-\alpha_{\text{slow}} \cdot t^*} &= e^{-\ln r / (r-1)} = r^{-1/(r-1)} \\
e^{-\alpha_{\text{fast}} \cdot t^*} &= e^{-r \cdot \ln r / (r-1)} = r^{-r/(r-1)}
\end{align}

Therefore:
\begin{equation}
\text{gap}_{\max} = \Delta \cdot \bigl(r^{-1/(r-1)} - r^{-r/(r-1)}\bigr)
\label{eq:gapmax_unfactored}
\end{equation}

This can be simplified by noting that $r^{-r/(r-1)} = r^{-1/(r-1)} \cdot r^{-1}$:
\begin{equation}
\text{gap}_{\max} = \Delta \cdot r^{-1/(r-1)} \cdot \left(1 - \frac{1}{r}\right)
\label{eq:gapmax}
\end{equation}

\subsection{Limiting behavior}

\textbf{Large timescale mismatch} ($r \to \infty$): As $r \to \infty$, $r^{-1/(r-1)} \to 1$ and $(1 - 1/r) \to 1$, so $\text{gap}_{\max} \to \Delta$. The peak gap approaches the full magnitude of the state change.

\textbf{Matched timescales} ($r \to 1^+$): Writing $r = 1 + \epsilon$ and expanding:
\begin{equation}
r^{-1/(r-1)} = (1 + \epsilon)^{-1/\epsilon} \to e^{-1} \quad \text{as } \epsilon \to 0
\end{equation}
while $(1 - 1/r) = \epsilon/(1+\epsilon) \to 0$. Thus $\text{gap}_{\max} \to 0$, confirming that the gap vanishes continuously as timescales converge.

\subsection{Cumulative gap}

The cumulative positive gap, which corresponds to the $\AUGpos$ metric used in the main text, integrates Eq.~\eqref{eq:gap} over the post-reversal period:
\begin{equation}
\AUGpos = \int_0^T \text{gap}(t)\,dt = \Delta \cdot \left(\frac{1}{\alpha_{\text{slow}}} - \frac{1}{\alpha_{\text{fast}}}\right) \cdot \bigl(1 - \text{residual}(T)\bigr)
\end{equation}
where the residual term vanishes as $T \to \infty$. In the infinite-horizon limit:
\begin{equation}
\AUGpos = \Delta \cdot \frac{r - 1}{\alpha_{\text{fast}}}
\end{equation}
This confirms that cumulative over-commitment scales linearly with timescale ratio and inversely with the fast learning rate.

\clearpage

\section{Supplementary Methods}

\subsection{Machine learning: extended experimental details}

\subsubsection{Dataset specifications}

The core experiments used five tabular benchmarks from UCI/OpenML:

\begin{table}[h]
\centering
\caption{Primary benchmark datasets.}
\begin{tabular}{lcccc}
\toprule
Dataset & Samples & Features & Classes & OpenML ID \\
\midrule
Ionosphere & 351 & 34 & 2 & 59 \\
Glass & 214 & 9 & 6 & 41 \\
Sonar & 208 & 60 & 2 & 40 \\
WDBC & 569 & 30 & 2 & 1510 \\
Vehicle & 846 & 18 & 4 & 54 \\
\bottomrule
\end{tabular}
\label{tab:datasets}
\end{table}

\noindent To inject noise, we flipped each training label to a uniformly random class with probability 0.4 (symmetric noise). Validation and test labels were left untouched.

The breadth validation (Extended Data Fig.~5) drew on 30 additional OpenML datasets, spanning 150--10{,}000 samples, 4--784 features, and 2--26 classes.

\subsubsection{Architecture details}

\textbf{Dense baseline.} Two hidden layers (128 and 64 units), ReLU, softmax output. Trained with Adam at $\text{lr} = 0.001$, batch size 32, for 50 epochs. Nothing fancy.

\textbf{Sparse-residual (ResEx).} The first hidden layer is replaced by a residual block with a fixed sparse branch:
\begin{equation}
h_1 = x + \alpha \cdot \sigma(W_\text{sparse} \cdot x)
\end{equation}
$W_\text{sparse}$ is an expander-graph adjacency matrix -- about 22\% dense, with $d = 3$ random neighbours per node. When $\alpha = 0$ the block is a pass-through; turning $\alpha$ up lets the sparse branch reshape the representation. The point is to have a single knob that smoothly dials memorization capacity.

\textbf{Controls.} Three architectures isolate individual ingredients:
\begin{itemize}
\item Dense+LS -- the dense baseline plus label smoothing ($\epsilon = 0.1$)
\item Dense+Residual -- dense with an identity shortcut but \emph{no} sparse bottleneck
\item Dense+StrongReg -- dense with aggressive $L_2$ ($\lambda = 0.01$)
\end{itemize}

\subsubsection{Memorization metric definitions}

The generalization gap at epoch $t$ is simply training accuracy minus validation accuracy:
\begin{equation}
\text{gap}(t) = \text{acc}_\text{train}(t) - \text{acc}_\text{val}(t)
\end{equation}

We then define two summary statistics. The first, $\AUGnorm$, captures cumulative memorization pressure:
\begin{equation}
\AUGnorm = \frac{1}{T} \sum_{t=1}^{T} \max(0, \text{gap}(t))
\end{equation}
where $T$ is the total number of epochs -- so $\AUGnorm$ is an average over the whole run, not just the tail.

The second, $\Tstar$, marks onset: the first epoch where the gap exceeds $\tau = 0.05$ for three epochs running. If that never happens, we set $\Tstar = T$ (right-censored). Both metrics are computed per seed and averaged.

\subsubsection{Alpha-grid causal intervention}

The logic is straightforward. If the sparse branch really controls memorization, then increasing $\alpha$ should push both $\AUGnorm$ and $\Tstar$ up monotonically -- but test accuracy should peak somewhere in the middle, because at high $\alpha$ the bottleneck starts hurting useful capacity too. We swept $\alpha \in \{0.1, 0.25, 0.5, 1.0\}$ across all five benchmarks (10 seeds each, 200 runs). Spearman $\rho$ between $\alpha$ and the group-median metric was 1.0 for both $\AUGnorm$ and $\Tstar$ ($p < 10^{-3}$): perfect monotonic control. Test accuracy peaked at $\alpha = 0.25$, exactly the kind of inverted-U you would expect from a capacity--memorization tradeoff.

\subsection{Human probabilistic reversal learning: extended methods}

\subsubsection{Dataset and participants}

We used trial-level data from Eckstein \textit{et al.} (2022)\cite{eckstein_reinforcement_2022}: 292 participants, ages 8--30, performing a probabilistic reversal learning task. The data are on OSF at \url{https://osf.io/7wuh4/}.

\subsubsection{Task structure}

On each trial, participants picked one of two options and got binary feedback (reward or no reward). One option paid off 75\% of the time, the other 25\%. These probabilities flipped 4--9 times per session (125--131 trials total). Because the ``good'' option still fails a quarter of the time, the effective noise rate works out to 17.1\% -- lower than the nominal 25\%, since the reward asymmetry partly disambiguates the contingency.

\subsubsection{Reversal-locked analysis}

Around each reversal we extracted a window from 8 trials before to 25 after. Within each window we computed two rolling accuracies (Gaussian-smoothed, $\sigma = 2$ trials):
\begin{itemize}
\item \textbf{Truth accuracy} -- fraction of choices matching the objectively better (post-reversal) option
\item \textbf{Feedback accuracy} -- fraction of choices that happened to receive positive feedback
\end{itemize}

\noindent The gap is the difference: feedback accuracy minus truth accuracy, trial by trial. We averaged across reversals within each subject first, then across subjects.

\subsubsection{$\AUGpos$ computation}

We wanted a single number capturing how much each subject over-committed to feedback after a reversal. $\AUGpos$ is the baseline-corrected area under the positive portion of the gap curve from trial 0 to trial 25:
\begin{equation}
\AUGpos = \frac{1}{25} \sum_{t=0}^{25} \max(0, \text{gap}(t) - \text{gap}_\text{baseline})
\end{equation}
The baseline ($\text{gap}_\text{baseline}$) is the average gap in the 8 pre-reversal trials, which removes any tonic offset that was already present before the contingency switched.

\subsection{Human reward/punishment learning with EEG: extended methods}

\subsubsection{Dataset and participants}

The dataset comes from Stolz, Endres \& Mueller (2022)\cite{stolz_reward_2022}: 26 healthy adults recorded with 32-channel EEG while doing separate reward and punishment probabilistic learning tasks (OpenNeuro ds004295). We dropped one subject whose EEG recording cut out mid-session, leaving $N = 25$ for behavior and $N = 23$ for RL model fits. Two more subjects lost too many epochs to artifact rejection in the reward condition (one in punishment), so the final neural samples are $N_\text{REW} = 24$ and $N_\text{PUN} = 25$.

\subsubsection{EEG preprocessing}

Raw \texttt{.set} files were read into MNE-Python (v1.6+). We bandpass-filtered at 1--40~Hz (zero-phase FIR), re-referenced to the average, cut feedback-locked epochs from $-200$ to $+600$~ms, subtracted the pre-stimulus baseline ($-200$ to 0~ms), and threw out any epoch where the voltage swing exceeded 150~$\mu$V. Nothing unusual here -- the pipeline is deliberately standard so the interesting part is what comes next.

\subsubsection{Neural features}

We picked three feedback-locked features, all of which have solid literatures linking them to outcome processing:

\begin{enumerate}
\item \textbf{Theta power (4--8~Hz) at FCz.} Morlet wavelets ($n_\text{cycles} = 4$), averaged 200--400~ms post-feedback. Midline frontal theta is the canonical EEG correlate of prediction-error signaling and feedback monitoring.

\item \textbf{Frontal beta (13--30~Hz).} Same time window, averaged across F3, F4, Fz, FC1, FC2 (Morlet, $n_\text{cycles} = 7$). Beta in this region has been tied to decision confidence and motor planning.

\item \textbf{P300 at Pz.} Mean ERP voltage from 250--450~ms -- a textbook marker of context updating after informative outcomes.
\end{enumerate}

All three were $z$-scored within subject and task before entering the decoder.

\subsubsection{Neural gap decoder}

We trained two logistic regressions per subject per task, using 3-fold cross-validation:
\begin{itemize}
\item A \textbf{feedback decoder} that predicts whether feedback was positive or negative from the three EEG features
\item A \textbf{truth decoder} that predicts whether the subject expected their choice to be correct before feedback (pre-feedback expectation), from the same features
\end{itemize}

On held-out folds, the two decoders produce probability estimates. The neural gap on trial $i$ is just the difference:
\begin{equation}
\text{neural\_gap}(i) = P(\text{feedback} = \text{positive} \mid \text{EEG}_i) - P(\text{expected correct} \mid \text{EEG}_i)
\end{equation}
Neural $\AUGpos$ is the mean of $\max(0, \text{neural\_gap}(i))$ across trials -- how much, on average, the brain over-represents feedback relative to expectation.

\subsubsection{Source localization}

To get a rough sense of where the signal lives on the scalp, we computed the positive-minus-negative feedback ERP difference in the 200--400~ms window at each electrode, averaging across the $N = 10$ subjects with enough trials in both valence conditions. We split electrodes into a frontal group (F3, F4, F7, F8, Fz, FC1, FC2, FC5, FC6, FCz, AF3, AF4, Fp1, Fp2) and a posterior group (P3, P4, P7, P8, Pz, PO3, PO4, O1, O2, Oz). Both tasks showed a frontal-greater-than-posterior pattern, consistent with the well-known role of medial frontal cortex in feedback evaluation.

\subsubsection{Cross-modal correlation}

At the subject level, we correlated each person's behavioral commitment (win-stay minus lose-shift) with their neural $\AUGpos$ using Spearman's $\rho$. The correlations were significant in both tasks: reward $\rho = 0.43$, $p = 0.036$; punishment $\rho = 0.42$, $p = 0.042$ (Fig.~5c). People whose brains over-represented feedback more strongly also showed more behavioral over-commitment -- the amplification is concordant across levels.

At the trial level, we smoothed both the neural gap and the behavioral gap with a 15-trial moving average, then correlated them (Pearson $r$) within each subject and averaged. Reward: $r = 0.67$, $p < 0.001$; punishment: $r = 0.53$, $p < 10^{-178}$.

\clearpage
\section{Supplementary Results}

\subsection{Complete statistical summary}

\begin{table}[h]
\centering
\caption{Summary of all primary statistical results across experiments.}
\small
\begin{tabular}{llclll}
\toprule
Analysis & $N$ & Measure & Statistic & $p$-value & Effect size \\
\midrule
\multicolumn{6}{l}{\textit{Human probabilistic reversal learning (Eckstein 2022)}} \\
PRL AUG$_\text{pos}$ & 292 & $0.049 \pm 0.032$ & $t(291) = 26.4$ & $1.1 \times 10^{-79}$ & $d = 1.55$ \\
PRL $T^*$ recovery & 292 & $9.9 \pm 13.9$ trials & 99.3\% & -- & -- \\
\midrule
\multicolumn{6}{l}{\textit{Human reward/punishment learning (Stolz 2022)}} \\
REW commitment & 25 & $0.567 \pm 0.171$ & $t(24) = 16.6$ & $5.9 \times 10^{-15}$ & $d = 3.32$ \\
PUN commitment & 25 & $0.628 \pm 0.160$ & $t(24) = 19.6$ & $1.4 \times 10^{-16}$ & $d = 3.92$ \\
REW win-stay & 25 & $0.799$ & -- & -- & -- \\
PUN win-stay & 25 & $0.829$ & -- & -- & -- \\
REW lose-shift & 25 & $0.232$ & -- & -- & -- \\
PUN lose-shift & 25 & $0.201$ & -- & -- & -- \\
\midrule
\multicolumn{6}{l}{\textit{Neural gap analysis (EEG)}} \\
REW neural AUG$_\text{pos}$ & 24 & $0.041 \pm 0.040$ & 100\% positive & -- & $d = 1.03$ \\
PUN neural AUG$_\text{pos}$ & 25 & $0.059 \pm 0.033$ & 100\% positive & -- & $d = 1.82$ \\
\midrule
\multicolumn{6}{l}{\textit{Cross-modal correlations}} \\
REW behav $\leftrightarrow$ neural & 24 & $\rho = 0.43$ & Spearman & $0.036$ & $\rho = 0.43$ \\
PUN behav $\leftrightarrow$ neural & 24 & $\rho = 0.42$ & Spearman & $0.042$ & $\rho = 0.42$ \\
\midrule
\multicolumn{6}{l}{\textit{ML alpha-grid causal intervention}} \\
AUG$_\text{norm}$ vs $\alpha$ & 200 runs & monotonic & $\rho = 1.0$ & $< 10^{-3}$ & -- \\
$T^*$ vs $\alpha$ & 200 runs & monotonic & $\rho = 1.0$ & $< 10^{-3}$ & -- \\
\bottomrule
\end{tabular}
\label{tab:all_stats}
\end{table}

\subsection{Reinforcement learning model parameters}

We fitted standard Rescorla--Wagner models to each participant in the Stolz (2022) dataset, with a single learning rate and softmax choice rule:

\begin{equation}
Q_{t+1}(a) = Q_t(a) + \alpha \cdot (r_t - Q_t(a))
\end{equation}
\begin{equation}
P(a_t = a) = \frac{\exp(\beta \cdot Q_t(a))}{\sum_{a'} \exp(\beta \cdot Q_t(a'))}
\end{equation}

Across $N = 23$ subjects, the fitted parameters were:
\begin{itemize}
\item REW: $\alpha = 0.439 \pm 0.314$, $\beta = 3.851 \pm 1.822$, decay $= 0.607 \pm 0.365$
\item PUN: $\alpha = 0.431 \pm 0.278$, $\beta = 4.137 \pm 1.222$, decay $= 0.582 \pm 0.374$
\end{itemize}

\noindent The reward and punishment parameters are strikingly similar, which fits with the parallel behavioral and neural signatures we observe across conditions.

\subsection{Positioning relative to noise-robust machine learning}

We are not trying to beat existing noise-robust training methods. There is already a large toolbox for that:

\begin{itemize}
\item \textbf{Loss correction}\cite{patrini_making_2017,natarajan_learning_2013} -- forward/backward correction, confident learning -- these adjust the loss to account for estimated noise transitions.
\item \textbf{Sample selection} -- Co-teaching, DivideMix, and relatives that identify and downweight suspicious examples during training.
\item \textbf{Implicit regularization} -- MixUp, AugMax, label smoothing -- approaches that reduce memorization as a side effect of data augmentation or soft targets.
\end{itemize}

\noindent Our question is different: not \textit{how} to fix label noise but \textit{when} and \textit{why} architectural changes alter memorization dynamics. The sparse-residual network is a probe, not a proposed solution. What matters for the paper is the monotonic, causal link between capacity and gap dynamics -- and the fact that the same measurement framework extends to human learning.

\clearpage
\section{Supplementary Figures}

\begin{figure}[ht]
\centering
\includegraphics[width=0.9\textwidth]{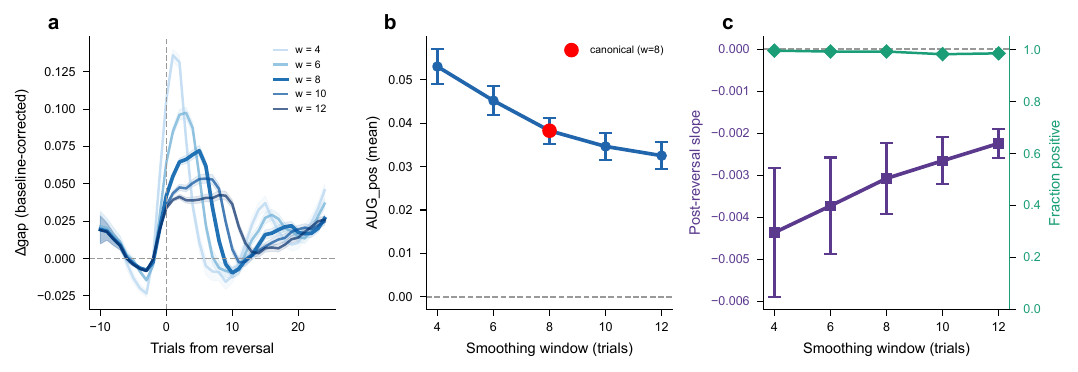}
\caption{\textbf{Supplementary Figure 1 $|$ PRL window sensitivity analysis.}
Robustness of the feedback--truth gap to rolling window size in the probabilistic reversal learning dataset ($N = 292$). Gap magnitude (AUG$_{\text{pos}}$) and temporal structure remain qualitatively consistent across window sizes from 4 to 16 trials, confirming that the observed dynamics are not artifacts of the chosen smoothing parameter. Main text uses 8-trial window (highlighted).}
\label{fig:supp1}
\end{figure}

\bibliography{references}